\documentclass[10pt, a4paper]{article}
\usepackage{lrec2022} 
\usepackage{multibib}
\newcites{languageresource}{Language Resources}
\usepackage{graphicx}
\usepackage{tabularx}
\usepackage{soul}
\usepackage{titlesec}
\titleformat{\section}{\normalfont\large\bfseries\center}{\thesection.}{1em}{}
\titleformat{\subsection}{\normalfont\SmallTitleFont\bfseries\raggedright}{\thesubsection.}{1em}{}
\titleformat{\subsubsection}{\normalfont\normalsize\bfseries\raggedright}{\thesubsubsection.}{1em}{}
\renewcommand\thesection{\arabic{section}}
\renewcommand\thesubsection{\thesection.\arabic{subsection}}
\renewcommand\thesubsubsection{\thesubsection.\arabic{subsubsection}}

\usepackage{epstopdf}
\usepackage[utf8]{inputenc}
\usepackage{hyperref}
\usepackage{xstring}
\usepackage{color}

\usepackage{microtype}
\usepackage{graphics}
\usepackage{multirow}
\usepackage{graphicx}
\usepackage{hyperref}
\usepackage{subfig}
\usepackage{amsfonts}
\usepackage{xcolor,colortbl}
\usepackage{comment}
\usepackage{hyperref}
\DeclareUnicodeCharacter{041C}{ö}
\title{A Multimodal Corpus for Emotion Recognition in Sarcasm}

\name{Anupama Ray, Shubham Mishra, Apoorva Nunna, Pushpak Bhattacharyya} 

\address{IBM Research India, Department of Computer Science and Engineering, IIT Bombay, India \\
         anupamar@in.ibm.com, shubham101mishra@gmail.com,
         \{apoorvanunna, pb\}@cse.iitb.ac.in\\}

\abstract{
While sentiment and emotion analysis have been studied extensively, the relationship between sarcasm and emotion has largely remained unexplored. A sarcastic expression may have a variety of underlying emotions. For example, “I love being ignored” belies sadness, while “my mobile is fabulous with a battery backup of only 15 minutes!” expresses frustration. Detecting the emotion behind a sarcastic expression is non-trivial yet an important task. We undertake the task of detecting the \textit{emotion in a sarcastic statement}, which to the best of our knowledge, is hitherto unexplored. We start with the recently released multimodal sarcasm detection dataset (MUStARD) pre-annotated with 9 emotions. We identify and correct 343 incorrect emotion labels (out of 690). We double the size of the dataset, label it with emotions along with \textit{valence} and \textit{arousal} which are important indicators of emotional intensity. Finally, we label each sarcastic utterance with one of the four sarcasm types-\textit{Propositional}, \textit{Embedded}, \textit{Likeprefixed} and \textit{Illocutionary}, with the goal of advancing sarcasm detection research. Exhaustive experimentation with multimodal (text, audio, and video) fusion models establishes a benchmark for exact emotion recognition in sarcasm and outperforms the state-of-art sarcasm detection. We release the dataset enriched with various annotations and the code for research purposes: \url{https://github.com/apoorva-nunna/MUStARD\_Plus\_Plus}
\\ \newline }

\begin{document}

\maketitleabstract
\section{Introduction} \label{sec:intro}

Emotion understanding leads to a deeper insight into the intent of the speaker and is key to generating the right response in conversational systems. Detecting emotions and sarcasm is crucial for all services involving human interactions, such as chatbots, e-commerce, e-tourism, and several other businesses. To be able to understand the user's intent, we started with the research problem on understanding the emotions that lead to the usage of sarcasm in a conversation. 
Sarcasm is a very sophisticated linguistic articulation where the surface meaning often 
stands in contrast to the underlying deeper meaning.
While this incongruity is the key element of sarcasm, the intent could be to appear humorous, ridicule someone, or to express contempt. Thus sarcasm is considered a very nuanced or intelligent language construct that poses several challenges to emotion recognition; for example, perceived emotion could be completely flipped due to the presence of sarcasm. Sarcasm often relies on verbal and non-verbal cues (pitch, tone, emphasis in speech, and body language in video). Even for humans, understanding the underlying emotion is challenging without the audio/video or the context of the conversation. However, researchers have worked on sarcasm detection on text modality with textual datasets (such as tweets \cite{oprea-magdy-2020-isarcasm}, Reddit short texts  \cite{DBLP:journals/corr/KhodakSV17}, dialogue \cite{oraby-etal-2016-creating} etc) for a decade. 

Recently we have seen multimodal datasets in the space of sarcasm detection, for example, image data from Twitter \cite{cai-etal-2019-multi}, code-mixed sarcasm and humor detection dataset \cite{msh}. \citelanguageresource{mustard} released a video dataset for sarcasm detection called MUStARD with 345 sarcastic videos and 345 non-sarcastic videos. \citelanguageresource{ACL2020PB} annotated MUStARD data with 9 emotion labels and sentiment (all sarcastic utterances having negative sentiment) and used emotion and sentiment to improve sarcasm detection. We started with this emotion-labeled variant of MUStARD provided by \citelanguageresource{ACL2020PB} to build a multiclass emotion recognizer on sarcastic utterances and observed several labeling errors while performing error analysis. During the annotation effort, we doubled the dataset by adding new utterances from similar sitcom genre series as in MUStARD while maintaining 50\% sarcastic and 50\% non-sarcastic videos. The affective dimensions of valence and arousal are commonly studied in the psychological and cognitive exploration of emotion \cite{EthicsSheet} and help in better understanding of emotion category and intensity. Thus the entire dataset is annotated with arousal and valence along with the perceived emotion of the speaker. While \textit{valence} indicates the extent to which the emotion is positive or negative, \textit{arousal} measures the intensity of the emotion associated \cite{cowie2003describing}. Finally, we also add sarcasm type as metadata which would help advance sarcasm detection research as well as give an understanding of what kind of information/modality is required to improve sarcasm detection. The four types of sarcasm are: \textit{Propositional}, \textit{Embedded}, \textit{Like-Prefixed} and \textit{Illocutionary} \cite{camp2012sarcasm}. \textbf{Propositional sarcasm} needs context information to be able to detect whether it's sarcasm or not. For example: ``your plan sounds fantastic!" may seem non-sarcastic if the context information is not present \cite{zvolenszky2012gricean}. \textbf{Embedded sarcasm} has an embedded incongruity within the utterance; thus, the text itself is sufficient to detect sarcasm. For example: ``It's so much fun working at 2 am at night". \textbf{Like-prefixed sarcasm} as the name suggests uses a like-phrase to show the incongruity of the argument being said, for example, ``Like you care" \cite{surveyJoshi}. \textbf{Illocutionary sarcasm} is a type of sarcasm that bears the sarcasm in the non-textual cues, and the text is often the opposite of the attitude captured in the audio or video modality. \cite{zvolenszky2012gricean} give an example of rolling eyes while saying "Yeah right" being a sarcastic sentence; although the text is sincere prosodic features in audio and eye movement in the video clearly show the sarcasm. 

The main contributions of this paper are:
\begin{itemize}
    \item An extended data resource which we call \textbf{MUStARD++} where we have doubled the existing MUStARD dataset and added labels for emotion, valence, arousal, and sarcasm-type information.
    \item Identify and correct labeling issues in emotion labels on MUStARD data presented in \citelanguageresource{ACL2020PB}.
    \item Exhaustive experimentation to benchmark multimodal fusion models for emotion detection in sarcasm.
\end{itemize} 

\begin{figure}[ht!]
\centering
\includegraphics[width=\columnwidth]{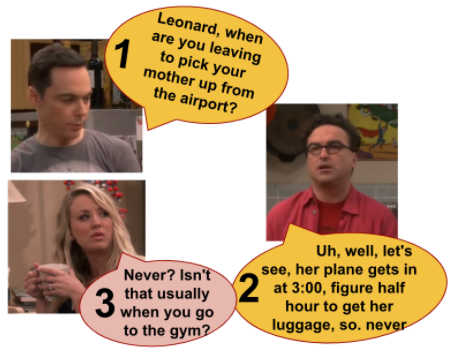}
\caption{Example to show that different explicit and implicit emotions in sarcasm. \textit{Explicit = Surprise}, and \textit{Implicit = Ridicule}; Sarcasm Type: \underline{Embedded}; Valence = 4; Arousal = 8}
\label{fig:intro-example}
\end{figure}

Figure \ref{fig:intro-example} is a sample in MUStARD++ with the labels and metadata information added to each video utterance. The text in the red bubble is the transcription of the sarcastic utterance, and the text in the yellow bubbles is the contextual sentences transcribed from the contextual video frames. The sarcasm is clearly evident just from the text modality (\textit{Embedded sarcasm}). This utterance is also an illustration of cases where the explicit emotion and implicit emotion of the speaker are different. This is common in sarcastic utterances where the speaker makes a sarcastic comment with either no expression or vocal changes but means quite the opposite. 

\section{Related Work} \label{sec:relted-work}
While there exist several studies on sentiment and emotion analysis, the relationship between emotion and sarcasm has been largely unaddressed. Most of the existing research has focused on the detection of sarcasm \cite{DBLP:journals/corr/JoshiBC16},\cite{SarcasmReviewBook}. Research studying the impact of sarcasm on sentiment analysis \cite{maynard2014cares} showed that sarcasm often has a negative sentiment, but the associated emotion(s) is important to frame the response and follow-up communication. 

\cite{Schifanella_2016} extended sarcasm detection to multimodal data (images and text) from social media and observed that visual features did boost the performance over the textual models. Along similar lines, \cite{multimodal_image} reported the improvement of the sarcasm detection task by using image data in addition to text. The dataset is curated from Instagram and the authors consider the image, text caption, and the transcript embedded within the image as multiple modalities. \cite{cai-etal-2019-multi} used text features, image features and image attributes as three modalities and proposed a multimodal hierarchical fusion model for sarcasm detection on tweets. 


MUStARD \citelanguageresource{mustard} is a subset of Multimodal Emotion Lines Dataset (MELD) \cite{MELD} and MELD is a multimodal extension of textual dataset EmotionLines \cite{EmotionLines}. MELD contains about 13,000 utterances from the TV series Friends, labeled with one of the seven emotions (anger, disgust, sadness, joy, neutral, surprise, and fear) and sentiment.
EmotionLines \cite{EmotionLines} and EmoryNLP \cite{emoryNLP} are textual datasets with conversational data, the former containing data from the TV show Friends and private Facebook messenger dialogues, while the latter was also curated from the series Friends.
Iemocap \cite{iemocap} is a well-known multimodal, dyadic dataset with 151 recorded videos annotated with categorical emotion labels, as well as dimensional labels such as valence, activation, and dominance. However, none of them have sarcastic utterances. \cite{msh} released a Hindi-English code-mixed dataset for the problem of Multimodal Sarcasm Detection and Humour Classification in a conversational dialog. They also propose an attention-based architecture named MSH-COMICS for enhanced utterance classification.
Along with categorical classification of basic emotions, seminal works \cite{russell-emotion,plutchick-emotion} also propose dimensional models of emotion (Ex: Valence, Arousal, Dominance), which could help in capturing the complicated nature of human emotions better. \cite{aff-wild} created a database of 298 videos (non-enacted, in-the-wild) and captured facial affect in their subjects in terms of valence arousal annotations ranging between -1 to +1. Similar work was undertaken in \cite{fb-va} where valence and arousal were annotated on a nine-point scale on Facebook data. They also release bag-of-words regression models trained on their data which outperform popular sentiment analysis lexicons in valence-arousal prediction tasks.

\section{Label Changes in MUStARD} \label{sec:label-issues}
\citelanguageresource{ACL2020PB} annotated the MUStARD dataset with emotions and sentiment and showed that emotion and sentiment labels could help sarcasm detection. Since our study mainly focuses on understanding the speaker's emotion leading to the use of sarcasm, we used their basic emotion annotation. After performing extensive experiments with all combinations of \textit{Video}, \textit{Text} and \textit{Audio} and several state-of-art models, we observed that most of the errors arise from the model predicting negative emotions when the true label is either \textit{Neutral} or \textit{Happy}. On detailed qualitative analysis, we observed that the labels for those sarcastic datapoints seemed intuitively incorrect. We built several models using each modality separately and also in combinations using different types of feature extractors and classifiers on the dataset. We flagged cases where majority of the models agreed with each other but disagreed with ground truth labels to obtain instances that needed re-annotation. We also grouped the error categories and the flagged cases to few distinct categories and observed that most of the errors are in sarcastic utterances.



Our analysis flagged 399 cases of disagreement with \citelanguageresource{ACL2020PB} out of the 690 video utterances. We initiated an unbiased manual labeling effort by a team of annotators on the entire dataset without giving them the labels from \citelanguageresource{ACL2020PB}. The re-annotation effort led to identifying 88 labeling issues in non-sarcastic and 255 labeling issues in sarcastic sentences.

A major chunk of errors (90 out of 345 sarcastic sentences) is in utterances previously labeled as Neutral. Literature shows that people resort to sarcasm in their utterances when they have negative intent or negative sentiment \cite{surveyJoshi}. 
In sarcasm, the explicit emotion can be positive, but the implied emotion/sentiment must have opposite polarity; hence it seemed unlikely for neutral or happy to appear in implicit emotion. 
Table \ref{tab:incorrect-label-1} shows an example of a label error wherein the utterance is marked as neutral for both explicit and implicit emotion. The sarcastic utterance (in gray) is expressed out of \textit{Disgust} and cannot be Neutral. Also, the audio and video clearly indicate that the speaker was overexcited to place the order before anyone else could speak. This particular utterance is an example of \textit{Propositional} sarcasm since we need the prior conversations to understand the sarcasm but the textual sentences are enough and doesn't need additional modalities for sarcasm detection. Additional modalities are however crucial for understanding the emotions for such cases. Our annotators felt that the cases labeled as neutral originally might have been difficult to annotate and thus were marked as neutral under majority voting.

\begin{table}[h]
\resizebox{\columnwidth}{!}{%
\begin{tabular}{|c|p{6cm}|c|c|c|c|}
\hline
\multirow{2}{*}{\textbf{Speaker}} &
  \multirow{2}{*}{\textbf{Utterance}} &
  \multicolumn{2}{c|}{\textbf{\citelanguageresource{ACL2020PB}}} &
  \multicolumn{2}{c|}{\textbf{New}} \\ \cline{3-6} 
 &
   &
  \textbf{Explicit} &
  \textbf{Implicit} &
  \textbf{Explicit} &
  \textbf{Implicit} \\ \hline
\begin{minipage}{.2\columnwidth}
      \fcolorbox{white}{white}{\includegraphics[width=\linewidth, height=\linewidth]{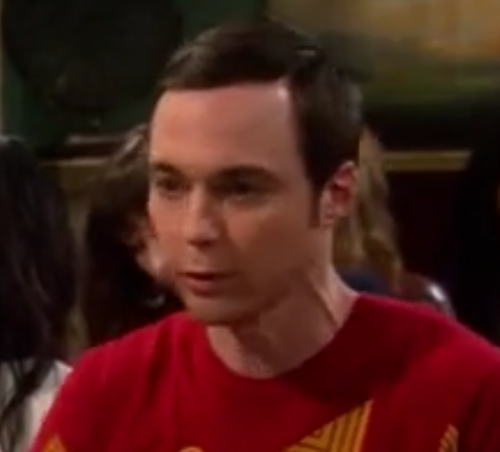}}
    \end{minipage} &
    The backwash into this glass is every pathogen that calls your mouth home, sweet home.  Not to mention the visitors who arrive on the dancing tongue of your subtropical girlfriend. &
  \cellcolor{lightgray}\multirow{4}{*}{Neu} &
  \cellcolor{lightgray}\multirow{4}{*}{Neu} &
  \cellcolor{lightgray}\multirow{4}{*}{Exi} &
  \cellcolor{lightgray}\multirow{4}{*}{Dis} \\ \cline{1-2}
\begin{minipage}{.2\columnwidth}
      \fcolorbox{white}{white}{\includegraphics[width=\linewidth, height=\linewidth]{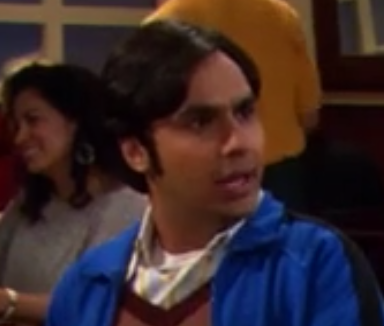}}
    \end{minipage} &
  Hey! That's my sister and my country  you're talking about. Leonard may  have defiled one, but I won't have you talking smack about the other. &\cellcolor{lightgray}
   &\cellcolor{lightgray}
   &\cellcolor{lightgray}
   &\cellcolor{lightgray}
   \\ \cline{1-2}
\begin{minipage}{.2\columnwidth}
      \fcolorbox{white}{white}{\includegraphics[width=\linewidth, height=\linewidth]{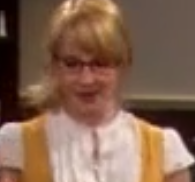}}
    \end{minipage} &
  You guys ready to order? & \cellcolor{lightgray}
   &\cellcolor{lightgray}
   &\cellcolor{lightgray}
   &\cellcolor{lightgray}
   \\ \cline{1-2}
\begin{minipage}{.2\columnwidth}
      \cellcolor{lightgray}\fcolorbox{lightgray}{lightgray}{\includegraphics[width=\linewidth, height=\linewidth]{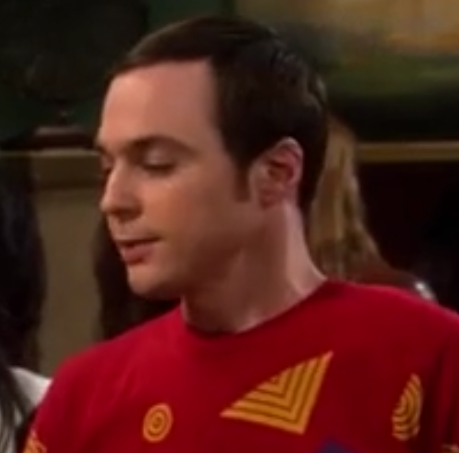}}
    \end{minipage} &\cellcolor{lightgray}
  \cellcolor{lightgray}  Yes, I'd like a seven-day course of penicillin, some, uh, syrup of ipecac-- to induce vomiting-- and a mint. &\cellcolor{lightgray}
   &\cellcolor{lightgray}
   &\cellcolor{lightgray}
   &\cellcolor{lightgray}
   \\ \hline
\end{tabular}%
}
\caption{Example of labeling error. Sarcastic utterance (in gray) and the 3 preceding utterances are context. Sarcasm type: \underline{Propositional} \scriptsize \{Neu- Neutral, Exi - Excitement, Dis - Disgust \} }
\label{tab:incorrect-label-1}
\end{table}

\begin{table}[h]
\centering
\resizebox{\columnwidth}{!}
{%
\tabcolsep=0.09cm
\begin{tabular}{|rll|lll|lll|lll|l|}
\cline{3-13}
\multicolumn{1}{l}{\multirow{3}{*}{}} &
  \multicolumn{1}{l|}{} &
  \multicolumn{5}{c|}{\textbf{Non-Sarcastic}} &
  \multicolumn{1}{l|}{} &
  \multicolumn{5}{c|}{\textbf{Sarcastic}} \\ \cline{3-7} \cline{9-13} 
\multicolumn{1}{l}{} &
  \multicolumn{1}{l|}{} &
  \multicolumn{2}{c|}{\textbf{Explicit}} &
  \multicolumn{1}{l|}{} &
  \multicolumn{2}{c|}{\textbf{Implicit}} &
  \multicolumn{1}{l|}{} &
  \multicolumn{2}{c|}{\textbf{Explicit}} &
  \multicolumn{1}{l|}{} &
  \multicolumn{2}{c|}{\textbf{Implicit}} \\ \cline{3-4} \cline{6-7} \cline{9-10} \cline{12-13} 
\multicolumn{1}{l}{} &
  \multicolumn{1}{l|}{} &
  \multicolumn{1}{c|}{\textbf{\textit{OLD}}} &
  \multicolumn{1}{c|}{\textbf{\textit{OUR}}} &
  \multicolumn{1}{l|}{} &
  \multicolumn{1}{c|}{\textbf{\textit{OLD}}} &
  \multicolumn{1}{c|}{\textbf{\textit{OUR}}} &
  \multicolumn{1}{l|}{} &
  \multicolumn{1}{c|}{\textbf{\textit{OLD}}} &
  \multicolumn{1}{c|}{\textbf{\textit{OUR}}} &
  \multicolumn{1}{l|}{} &
  \multicolumn{1}{c|}{\textbf{\textit{OLD}}} &
  \multicolumn{1}{c|}{\textbf{\textit{OUR}}} \\ \cline{1-4} \cline{6-7} \cline{9-10} \cline{12-13} 
\textbf{Anger}      &  & 28  & 28 &  & 35  & 28 &  & 26  & 3   &  & 62 & {\color[HTML]{006400} \textbf{68}} \\
\textbf{Excitement} &  & 15  & 32 &  & 15  & 31 &  & 15  & 41  &  & {\color[HTML]{B22222} \textbf{2}}  & 0  \\
\textbf{Fear}       &  & 6   & 8  &  & 10  & 9  &  & 0   & 0   &  & 4  & 0  \\
\textbf{Sad}        &  & 62  & 61 &  & 68  & 62 &  & 56  & 47  &  & 53 & {\color[HTML]{006400} \textbf{18}} \\
\textbf{Surprise}   &  & 20  & 19 &  & 18  & 18 &  & 15  & 21  &  & {\color[HTML]{B22222} \textbf{11}} & 0  \\
\textbf{Frustrated} &  & 13  & 27 &  & 17  & 28 &  & 10  & 3   &  & 40 & {\color[HTML]{006400} \textbf{88}} \\
\textbf{Happy}      &  & 92  & 82 &  & 81  & 80 &  & 114 & 85  &  & {\color[HTML]{B22222} \textbf{62}} & 0  \\
\textbf{Neutral}    &  & 115 & 75 &  & 108 & 75 &  & 113 & 144 &  & {\color[HTML]{B22222} \textbf{90}} & 0  \\
\textbf{Disgust}    &  & 6   & 13 &  & 7   & 14 &  & 4   & 0   &  & 32 & {\color[HTML]{006400} \textbf{81}} \\
\textbf{Ridicule{\scriptsize *}}   &  & -   & 0  &  & -   & 0  &  & -   & 0   &  & -  & {\color[HTML]{006400} \textbf{89}} \\ \cline{1-13} 
\end{tabular}%
}
\caption{Emotion distribution comparison in MUStARD between earlier and updated labels.(\textit{OLD} is \protect\citelanguageresource{ACL2020PB} annotations and \textit{OUR} is proposed annotations.)\textit{\scriptsize *Ridicule in introduced in new annotations}}
\label{tab:emotion distribution}
\end{table}

Table \ref{tab:emotion distribution} shows the number of label changes that were done on original MUStARD dataset. As seen in Table \ref{tab:emotion distribution}, most labeling errors are in sarcastic utterances due to the challenges sarcasm adds. 
There were 62 sarcastic utterances which were labeled with happy as implicit emotion and 114 sarcastic sentences had happy as explicit emotion. While happy can be an explicit emotion, our annotators suggested that the correct intent for such sarcastic utterances should be ridicule or mockery as these shows belong to the genre of situational comedy (sit-com), wherein characters use sarcasm to ridicule their friends while demonstrating happiness explicitly. Thus we introduced a new label \textit{Ridicule} and allowed annotators to label as per these labels. 

\section{Dataset} \label{sec:dataset}
Towards understanding emotions in sarcasm, we had two main challenges: difficulty in getting multimodal sarcastic data, and challenges of annotating the perceived emotion of a speaker for every sarcastic utterance. While 10,000 non-sarcastic videos could be gathered in MELD \cite{MELD}, only 345 out of them were sarcastic \cite{mustard} which stands proof to the difficulty in finding multimodal sarcasm data. In this work, we doubled the size of this dataset, by carefully adding sarcastic videos from similar genre, annotate it with sarcasm presence or absence, as well as emotions, arousal and valence. We also point out that to improve sarcasm detection, it is important to understand the type of sarcasm present and thus annotate each video with the sarcasm types - \textit{Propositional, Illocutionary, Like-Prefixed} and \textit{Embedded}. 

\subsection{Data collection}
In \cite{mustard}, authors collected videos from sit-com TV shows: Friends, Big Bang Theory (seasons 1–8), The Golden Girls, and Burnistoun (also referred to as Sarcasmaholics Anonymous). We collected all videos from The Big Bang Theory season 9-12, out of which we could only get 216 sarcastic video utterances. We considered another series of similar genre called The Silicon Valley\footnote{\url{https://en.wikipedia.org/wiki/Silicon_Valley_(TV_series)}} which has 6 seasons of 53 episodes. Out of the 53 episodes, we could only find 41 video utterances that are sarcastic. Although all such videos have humor, not all are sarcastic, thus needing careful observation while selecting and manual annotation. We added equal number of non-sarcastic videos with context to create balanced sarcasm detection dataset. While in non-sarcastic sentences a speaker might have only one emotion, sarcasm due to its incongruous nature, exhibits an extrinsic and an intrinsic emotion. 


\subsection{Annotation Protocol and Observations}
\begin{figure}[ht!]
\centering
        \includegraphics[totalheight=5cm]{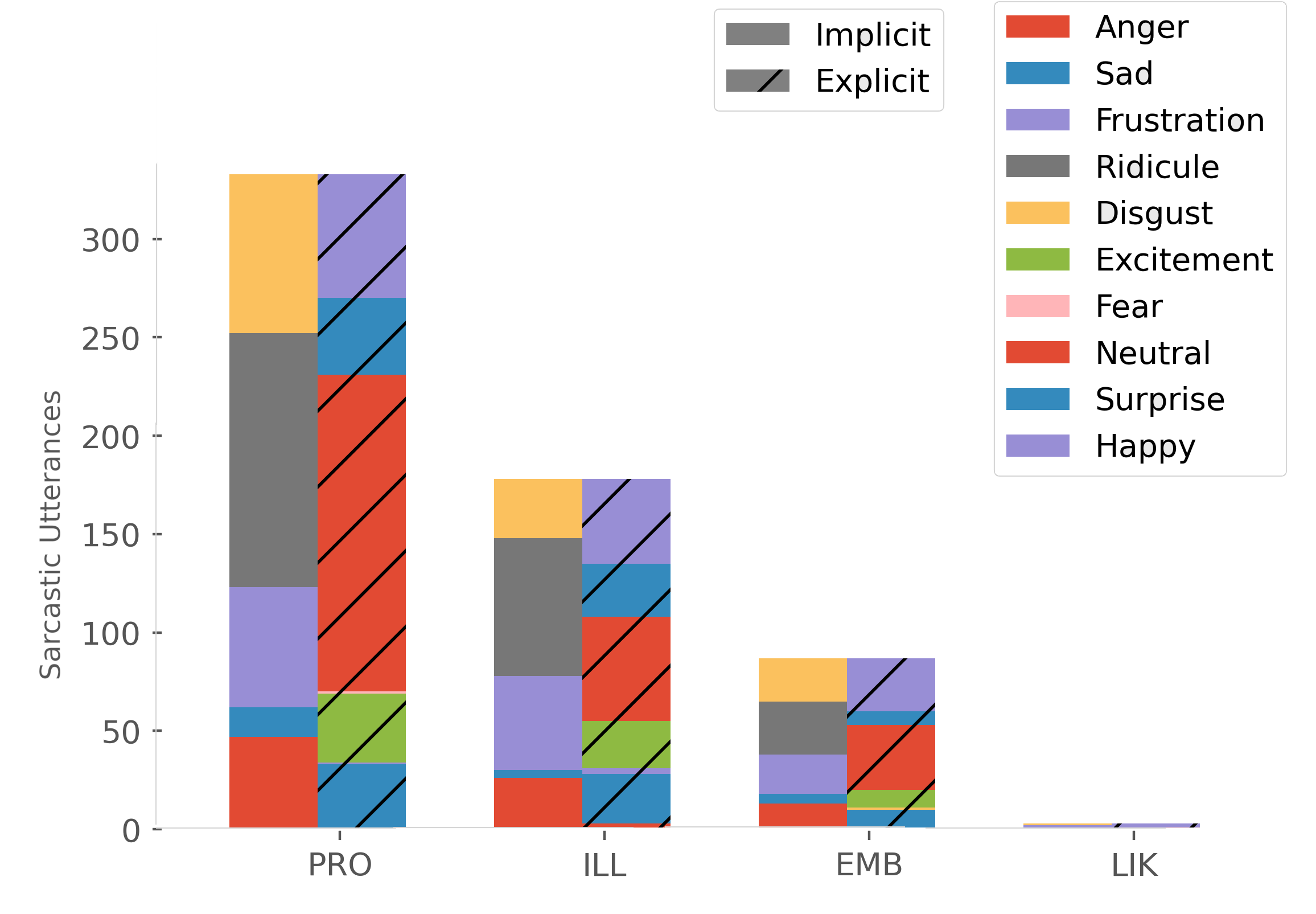}
        \captionsetup{justification=centering}
    \caption{Distribution of Emotion over Sarcasm Type in Proposed Dataset}
    \label{fig:implicit-emotion-sat-type}
\end{figure}

We employ seven annotators proficient in English. While one annotator has a Ph.D. in Humanities, two of them were linguists; others were engineering graduates. They were selected from a pool of annotators due to their prior working experience in the field of Sentiment and Emotion Analysis and understanding of emotion and sarcasm. We have four male and three female annotators and all annotators were in the age group of 18-30. They were provided the detailed instructions on the annotation protocol before beginning the annotation process with examples of each type of sarcasm. In the first round of manual annotation, we gave our annotators the original MUStARD dataset for emotion annotation and asked them to put their labels for extrinsic and intrinsic emotion of the speaker without access to the emotion labels provided by \citelanguageresource{ACL2020PB}. Instead of annotating only those videos where we observed the incorrect labels, we decided to annotate all videos of existing dataset as well as our newly collected video utterances. Annotators had access to full videos for annotation but were instructed to start with only text transcription of the utterance, then proceed with text transcriptions of contextual frames and finally watch the utterance and context video. Following this annotation protocol is especially important to be able to correctly classify the sarcasm type. For example, text to be observed in isolation for Embedded and Like-Prefixed sarcasm, transcript of utterance and context should be observed in isolation for Propositional sarcasm, and the whole video to be considered for Illocutionary sarcasm. Out of the 601 sarcastic videos, we have 333 Propositional \textbf{( 55.4\%)}, 178 Illocutionary \textbf{29.6\%}, 87 Embedded \textbf{14.4\%} and 3 Like-prefixed \textbf{0.4\%} sarcastic videos in this proposed dataset. Figure \ref{fig:implicit-emotion-sat-type} shows a breakdown of emotion distribution per type of sarcasm present. 

Below are some examples of labeling issues: utterances with different extrinsic and intrinsic emotions and the emotion label changes introduced for more clarity. We also provide insights that help understand the challenges in emotion recognition in sarcasm. These were previously annotated as Neutral explicit emotions, which were changed to Ridicule 

\begin{table}[ht!]
\resizebox{\columnwidth}{!}{%
\begin{tabular}{|c|c|p{7cm}|c|c|c|}
\hline
Sr. No &
  \multicolumn{1}{c|}{Speaker} &
  \multicolumn{1}{c|}{Utterance} &
  \multicolumn{1}{c|}{\textbf{E}} &
  \multicolumn{1}{c|}{\textbf{I}} &
  \multicolumn{1}{c|}{\textbf{TYPE}} \\ \hline
\multirow{2}{*}{1} &
  \multicolumn{1}{c|}{\begin{minipage}{.2\columnwidth}
      \fcolorbox{white}{white}{\includegraphics[width=\linewidth, height=\linewidth]{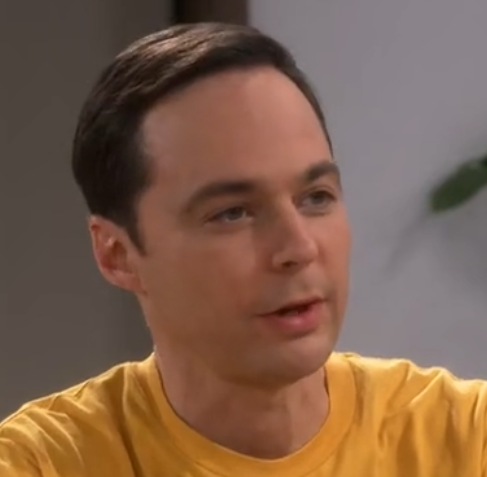}}
    \end{minipage}} &
  \multicolumn{1}{p{7cm}|}{I think Howard hurting my feelings has in some ways made me a better person. } &
  \multicolumn{1}{c|}{\multirow{-2}{*}{\cellcolor{lightgray} Sur}} &
  \multicolumn{1}{c|}{\multirow{-2}{*}{\cellcolor{lightgray} Rid}} &
  \multicolumn{1}{c|}{\multirow{-2}{*}{\cellcolor{lightgray} EMB}} \\ \cline{2-3}
 &
  \multicolumn{1}{c|}{\cellcolor{lightgray}\begin{minipage}{.2\columnwidth}
      \fcolorbox{lightgray}{lightgray}{\includegraphics[width=\linewidth, height=\linewidth]{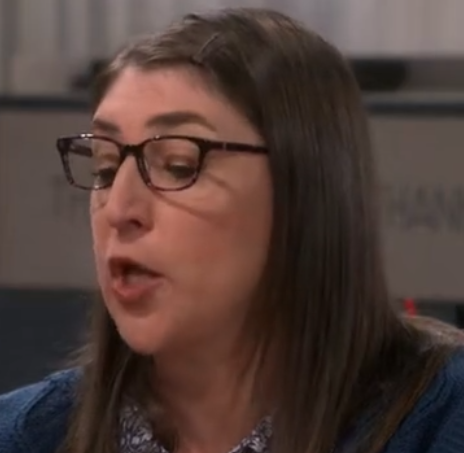}}
    \end{minipage}} &
  \multicolumn{1}{p{7cm}|}{\cellcolor{lightgray}Hmm. Look at you, improving on perfection. How so?} &
  \multicolumn{1}{c|}{\cellcolor{lightgray} } &
  \multicolumn{1}{c|}{\cellcolor{lightgray} } &
  \multicolumn{1}{c|}{\cellcolor{lightgray} } \\ \hline
\multicolumn{5}{|l|}{} \\ \hline
\multirow{3}{*}{2} &
  \begin{minipage}{.2\columnwidth}
      \fcolorbox{white}{white}{\includegraphics[width=\linewidth, height=\linewidth]{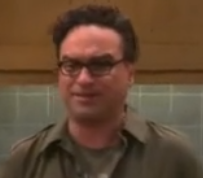}}
    \end{minipage} &
  Wow. That's a lot of luggage for a weekend. &\cellcolor{lightgray}
  \multirow{3}{*}{Neu} &\cellcolor{lightgray}
  \multirow{3}{*}{Rid} &\cellcolor{lightgray}
  \multirow{3}{*}{PRO} \\ \cline{2-3}
 &
  \begin{minipage}{.2\columnwidth}
      \fcolorbox{white}{white}{\includegraphics[width=\linewidth, height=\linewidth]{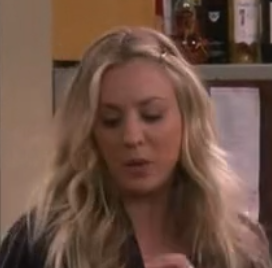}}
    \end{minipage} &
  (groans) I know. I didn't know what to wear, so I brought a few options. &\cellcolor{lightgray}
   &\cellcolor{lightgray} &\cellcolor{lightgray}
   \\ \cline{2-3}
 &
  \cellcolor{lightgray}\begin{minipage}{.2\columnwidth}
      \fcolorbox{lightgray}{lightgray}{\includegraphics[width=\linewidth, height=\linewidth]{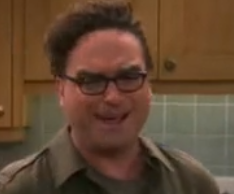}}
    \end{minipage} &\cellcolor{lightgray}
  Was one of the options the option to never come back? &\cellcolor{lightgray}
   &\cellcolor{lightgray}
    &\cellcolor{lightgray}
   \\ \hline
\multicolumn{5}{|l|}{} \\ \hline
\multirow{3}{*}{3} &
  \begin{minipage}{.2\columnwidth}
      \fcolorbox{white}{white}{\includegraphics[width=\linewidth, height=\linewidth]{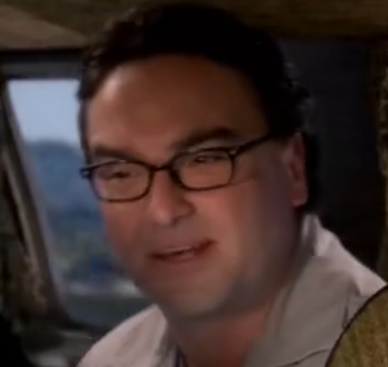}}
    \end{minipage} &
  So, are you gonna give us a clue where we're headed? &\cellcolor{lightgray}
  \multirow{3}{*}{Neu} &\cellcolor{lightgray}
  \multirow{3}{*}{Rid} &\cellcolor{lightgray}
  \multirow{3}{*}{PRO} \\ \cline{2-3}
 &
  \begin{minipage}{.2\columnwidth}
      \fcolorbox{white}{white}{\includegraphics[width=\linewidth, height=\linewidth]{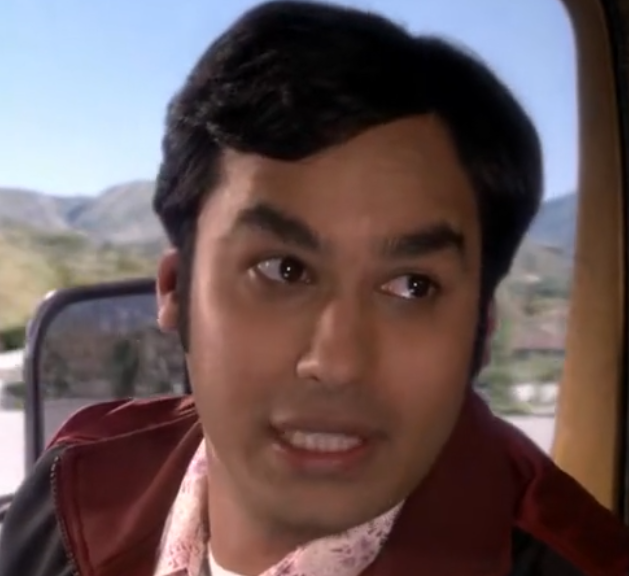}}
    \end{minipage} &
 Uh, okay, let's see...  They've got spicy food and  there's a chance you'll get diarrhea. &\cellcolor{lightgray} &\cellcolor{lightgray} &\cellcolor{lightgray}
   \\ \cline{2-3}
 &
  \cellcolor{lightgray}\begin{minipage}{.2\columnwidth}
      \fcolorbox{lightgray}{lightgray}{\includegraphics[width=\linewidth, height=\linewidth]{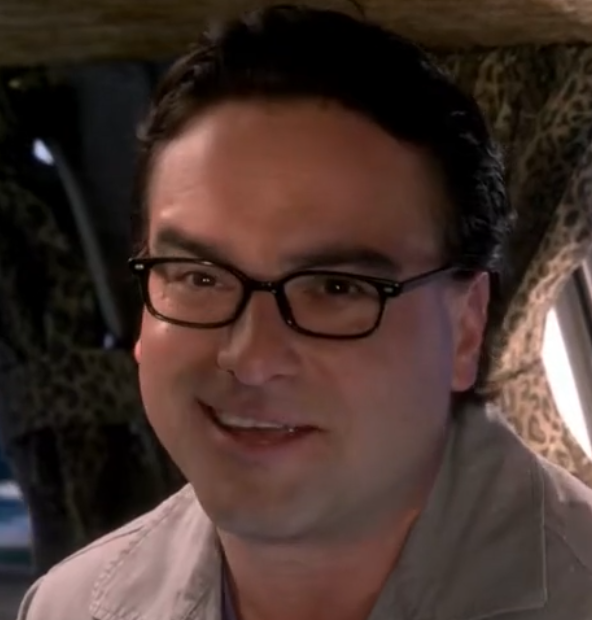}}
    \end{minipage} &\cellcolor{lightgray}
  India &\cellcolor{lightgray}
   &\cellcolor{lightgray} &\cellcolor{lightgray}
   \\ \hline
\end{tabular}%
}
\captionsetup{justification=centering}
\caption{Some examples of ridicule}
\label{tab:ridicule examples}
\end{table}

\begin{table}[ht!]
\resizebox{\columnwidth}{!}{%
\begin{tabular}{|c|p{6cm}|c|c|c|c|}
\hline
\multirow{2}{*}{\textbf{Speaker}} &
  \multirow{2}{*}{\textbf{Utterance}} &
  \multicolumn{2}{c|}{\textbf{\citelanguageresource{ACL2020PB}}} &
  \multicolumn{2}{c|}{\textbf{New}} \\ \cline{3-6} 
            &       &  
  \textbf{Explicit} &
  \textbf{Implicit} &
  \textbf{Explicit} &
  \textbf{Implicit}  \\ \hline 
\begin{minipage}{.2\columnwidth}
      \fcolorbox{white}{white}{\includegraphics[width=\linewidth, height=\linewidth]{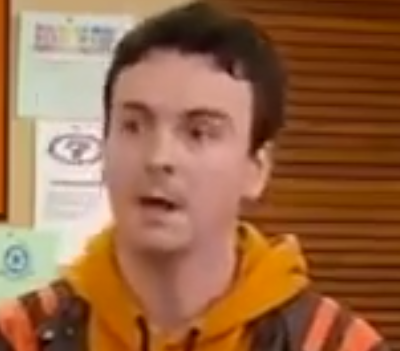}}
    \end{minipage} &
  My name's Scott and I am a sarcasmaholic. &\cellcolor{lightgray}
  \multirow{3}{*}{Sur} &\cellcolor{lightgray}
  \multirow{3}{*}{Sur} &\cellcolor{lightgray}
  \multirow{3}{*}{Sur} &\cellcolor{lightgray}
  \multirow{3}{*}{Rid} \\ \cline{1-2}
\begin{minipage}{.2\columnwidth}
      \fcolorbox{white}{white}{\includegraphics[width=\linewidth, height=\linewidth]{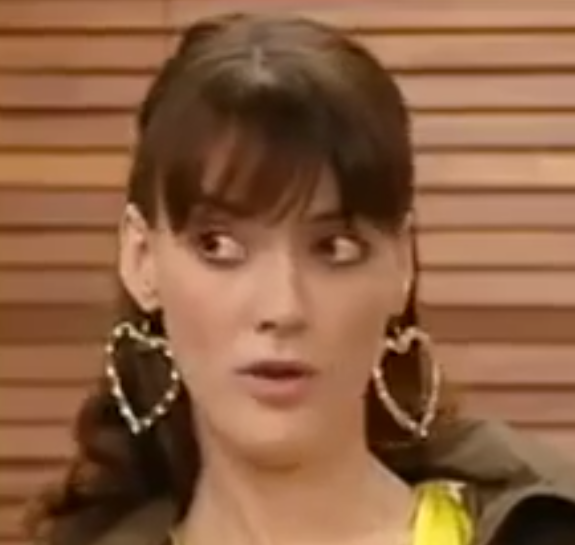}}
    \end{minipage} & Nooo. &\cellcolor{lightgray}&\cellcolor{lightgray}&\cellcolor{lightgray}&\cellcolor{lightgray}\\ \cline{1-2}
\cellcolor{lightgray}\begin{minipage}{.2\columnwidth}
      \fcolorbox{lightgray}{lightgray}{\includegraphics[width=\linewidth, height=\linewidth]{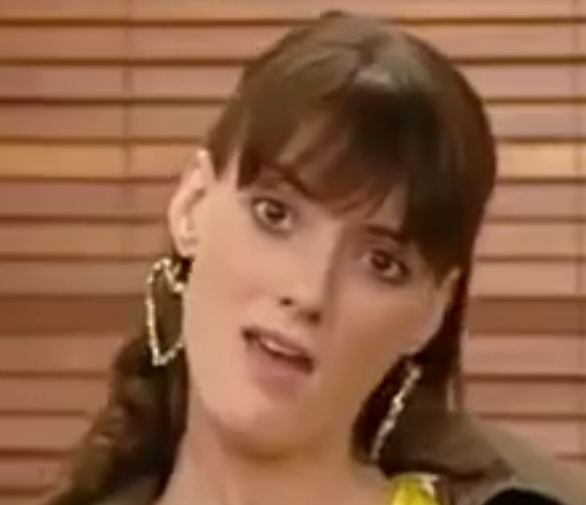}}
    \end{minipage} &
  \cellcolor{lightgray} We thought you were just here for the company. He is a sarcasmoholic Stewart.&\cellcolor{lightgray}
   &\cellcolor{lightgray}
   &\cellcolor{lightgray}
   &\cellcolor{lightgray}
   \\ \hline
\end{tabular}%
}

\caption{Example of Incorrect Labeling: Text transcription might suggest \textit{Surprise} as the emotion, however video makes it apparent that the intent is to \underline{ridicule}. Sarcasm Type: \underline{Illocutionary}\scriptsize  \{Sur- Surprise, Rid-Ridicule \}}
\label{tab:incorrect-label-3}
\end{table}


Initially we kept the emotion labels the same as \citelanguageresource{ACL2020PB} (i.e. \textit{Anger, Excitement, Fear, Sad, Surprise, Frustrated, Happy, Neutral, Disgust}), but after annotating 25 to 30\% videos, annotators suggested for a label which is in between frustration and disgust, and close to mockery. We looked at some examples (one of them is mentioned in Figure-\ref{fig:intro-example}) and named this category \textbf{Ridicule}. According to Plutchik's wheel of emotion \cite{plutchick-emotion} contempt is a higher-order emotion composed of the two basic emotions - anger and disgust. Since in this genre of situational comedy, the most likely emotion is frustration and disgust, and not anger, we cannot directly call all such instances contempt. Also, the intent of the speaker of the sarcastic utterance is predominantly to mock or ridicule; thus, we called this category Ridicule. Also, Ridicule is not a basic emotion present in the gold standard emotion scales \cite{plutchick-emotion,ekman}, and we call it just a category in our labeling scheme and not a basic emotion.\\

\begin{table}[ht!]
\resizebox{\columnwidth}{!}{%
\begin{tabular}{|l|c|c|c|c|c|c|c|c|c|c|}
\hline
                 & \textbf{An} & \textbf{Ex} & \textbf{Fe} & \textbf{Sa} & \textbf{Sp} & \textbf{Fs} & \textbf{Hp} & \textbf{Neu} & \textbf{Dis} & \textbf{Ri} \\ \hline
\textbf{Explicit} & 4                    & 68                   & 1                    & 67                   & 73                   & 4                    & 135                  & 248                   & 1                     & 0                    \\
\textbf{Implicit} & 87                   & 0                    & 0                    & 24                   & 0                    & 130                  & 0                    & 0                     & 134                   & 226                  \\ \hline
\end{tabular}%
}
\caption{Emotion distribution in Sarcastic Utterances of MUStARD++. \footnotesize \{An-Anger, Ex- Excitement, Fe-Fear, Sa-Sad, Sp-Surprise, Fs-Frustrated, Hp-Happy Neu-Neutral, Dis-Disgust, Ri-Ridicule\}}
\label{tab:emo-dis-new}
\end{table}

During re-annotation, 345 of the sarcastic videos from MUStARD are annotated as sarcastic, implying that the annotators have good understanding of sarcasm. Also, out of 264 new videos, 8 videos are moved to non-sarcastic because at least one annotator annotated it as non-sarcastic, making the total number of sarcastic videos in MUStARD++ 601. We increased the number of non-sarcastic videos to 601 in the extended dataset. The final label was chosen via majority voting. The overall inter-annotator agreement was calculated with a Kappa score of 0.595, which is comparable with the Kappa score of 0.5877 of original MUStARD annotations. Table \ref{tab:emo-dis-new} shows the explicit and implicit emotion distribution of the extended dataset. 
Additionally, annotators were instructed to rate each utterance with a valence and an arousal rating ranging from 1 to 9, with 1 being extremely low, 9 being extremely high and 5 being neutral. Pair-wise Quadratic Cohen Kappa was used to evaluate the inter-annotator agreement, and the average agreement was found to be 0.638 for valence and 0.689 for arousal.



\begin{table}[]
\resizebox{\columnwidth}{!}{%
\begin{tabular}{lccc}
Statistics              & \multicolumn{1}{l}{} & Utterance & Context \\ \hline
Unique words            & -                    & 2695      & 4891    \\
Avg. length (tokens)    & -                    & 11.3        & 7.8       \\
Max  length (tokens)    & -                    & 65        & 47      \\
Avg. duration (seconds) & -                    & 4.19      & 8.69   \\ \hline
\end{tabular}%
}
\caption{ Dataset statistics by utterance and context}
\label{tab:u-and-c-stats}
\end{table}

\section{Experiments} \label{sec:experiments}
Although our primary goal is to detect emotions in sarcasm, we also benchmark the proposed extended dataset for sarcasm detection and valence-arousal prediction. We perform various experiments examining different modalities independently and in various combinations. We analyze the impact of context and speaker information in each of the models. In the speaker dependent setup we passed the speaker information as a one-hot vector along with the utterance. When no such speaker information is passed, that method is being referred to as speaker-independent method. This was done as we observed that there are specific characters in each series who pass most of the sarcastic comments. Thus we wanted to see if the models benefit from the speaker information or not.

\subsection{Preprocessing and Feature Extraction}
Owing to the presence of multiple modalities, the features from text, audio and video were separately extracted and fused appropriately to act as input into our model. We discuss our feature extraction methods in detail below.

\textbf{Textual Modality}\\
In order to extract features from transcript (context and utterance), we tried using different transformer models such as BERT \cite{bert},BART \cite{BART}, RoBERTa \cite{RoBERTa} and T5\cite{T5}. BART performed slightly better in all experiments (only text, as well as in combination with audio and video) over BERT, RoBERTa and T5 models, thus we continued with BART-large representations for text. BART provides a feature vector representation $\mathbf{x_\mathit{t} \in \mathbb{R}^{\mathit{d_t}}}$ for every instance $\mathbf{x}$. We encode the text using BART Large model with ${{\mathit{d_t = 1024}}}$ and use the mean of the last four transformer layer representations to get a unique embedding representation for both utterance and context.

\textbf{Audio Modality}\\
We extract low-level features from the audio data stream for each utterance in the dataset to take advantage of information from the audio modality. We sampled the audio signal at 22.5KHz. Since the audio has background noise and canned laughter, we used vocal-separation method to process it \footnote{\url{https://librosa.org/doc/main/auto_examples/plot_vocal_separation.html}}. We extracted three low-level features: Mel Frequency Cepstral Coefficients (MFCC), Mel spectrogram (using Librosa library\cite{brian_mcfee_2022_6097378}), and prosodic features using OpenSMILE \footnote{\url{https://audeering.github.io/opensmile/}}.

We split the audio signal into equal length segments to maintain consistent feature representation in all instances. Since the audio signal length varies for different utterances, this segmentation helps in keeping vector size constant across dataset. For each segment we extract MFCC, Mel spectrogram and prosodic features of size $\mathit{d_m, d_s, d_p}$ respectively. Then we take the average across segments to get the final feature vector. Here $\mathit{d_m = 128, d_s = 128, d_p = 35}$, so our audio feature vector is of size $\mathit{d_a = 291}$.
We had also experimented with self-supervised speech encoders \cite{Pase2019}. However due to the very small number of sarcastic utterances, such models are unable to learn and thus we decided to stick with our low-level audio features.\\
\textbf{ Video Modality}\\
In order to extract visual features from the videos, we have used a pool5 layer of pre-trained ResNet-152 \cite{ResNet} image classification model. To improve the video representation and reduce noise, we extracted the key frames to be passed to ResNet-152, instead of feeding in information from all of the frames. Key frame extraction is widely used in the vision community and is defined as the frames that form the most appropriate summary of a given video \cite{jadon2019video}. We used an open source tool called Katna\footnote{\url{https://katna.readthedocs.io/en/latest/}}, to perform key-frame extraction. For final feature vectors we average the vectors of each key frame of an instance (context and utterance) extracted from ResNet-152. The size of final video feature representation in $\mathit{d_v = 2048}$.

\subsection{Experimental Setup}
We perform multi-class emotion classification experiments using the features extracted. Since we have three modalities, context and speaker information, we perform several ablation studies to understand the impact of presence or absence of each of these aspects. 
For the multi-modal fusion, we use collaborative gating architecture introduced in \cite{liu2020use}, with the only difference that not all the input embeddings need to be pretrained. First we calculate projection $\mathit{\Psi^{(i)}(V)}$ where  $\mathit{i \in \{t,tc,a,ac,v,vc\}}$  and  $\mathit{t,a,v,c}$ are \textit{text, audio, video} and \textit{ corresponding context}. The collaborative gating module implements two tasks: first, we find the attention vector prediction for three main input projections referred to as projection embeddings (i.e for our utterances in three modalities) $\mathit{T = \{ T^{t}(V),T^{a}(V),T^{v}(V)\}}$.
\begin{equation}
    \mathit{T^{(i)}(V) = h_{\phi }(\sum_{j\neq i}g_{\theta }(\Psi^{(i)}(V),\Psi^{(j)}(V)) )}
\end{equation}
where functions $\mathit{h_{\phi}}$ and $\mathit{g_{\theta}}$ are used to model the pairwise relationship between projection $\mathit{\Psi^{(i)}}$ and $\mathit{\Psi^{(j)}}$. Also $\mathit{i \in \{t,a,v\}}$ and $\mathit{j \in \{t,tc,a,ac,v,vc\}}$.

Then we perform expert response modulation using the attention vector prediction calculated. For response modulation of each modality projection we perform- 
\begin{equation}
    \mathit{\Psi^{(i)}(V) = \Psi^{(i)}(V)  \circ \sigma (T^{(i)}(V)) }
\end{equation}
where $\sigma$ is an element-wise sigmoid activation and $\circ$ is the element-wise multiplication (Hadamard product).  All the modulated projections are concatenated and passed to fully connected linear layers (ReLU) followed by a softmax layer to predict target class probability distribution. Cross entropy loss is used for all classification experiments. 
\\For completeness in bench-marking our data, we also perform Majority sampling (assigns the emotion class with majority examples as all samples), Random Sampling (predictions are sampled equally throughout the test set using this baseline). We also perform one-vs-rest experiments for each emotion which contain a sigmoid layer instead of a softmax layer as classification head. The results of these baselines are reported in the supplementary material section \ref{'sec:appendix'}. In addition to emotion classification, similar set up was used to study the performance of sarcasm detection on our dataset which was treated as a binary classification problem. Furthermore, we build a regression model for valence-arousal prediction based on the same architecture explained above, with the only difference of ReLU layer replacing the classification head of classification models and the loss function being the smooth L1 loss. For all training we perform hyper-parameter search with dropout in range of [0.2,0.3,0.4], learning rate in [0.001,0.0001], batch size [32,64,128], shared embedding size [2048, 1024] and projection embedding size [1024,256].

\begin{table*}[ht!]
\centering
\resizebox{0.8\textwidth}{!}{%
\begin{tabular}{|c|ccc|c|ccc|}
\hline
Methods & \multicolumn{3}{c|}{\textit{\textbf{Speaker Independent}}}   &                                      & \multicolumn{3}{c|}{\textit{\textbf{Speaker Dependent}}}   \\ \cline{2-4} \cline{6-8} 
\multirow{-2}{*}{}   & \textbf{P} & \textbf{R}   & \textbf{F1}  &   & \textbf{P}  & \textbf{R}  & \textbf{F1}  \\ \cline{1-4} \cline{6-8} 
\textbf{\citelanguageresource{mustard}}                                    & 64.7                                               & 62.9                                              & 63.1                                               &                                      & 72.1                                               & 71.7                                              & 71.8                                               \\
\textbf{\citelanguageresource{ACL2020PB}}                                        & 69.53                                     & 66.0                                              & 65.9                                               &                                      & 73.40                                     & 72.75                                             & 72.57                                              \\
\textbf{Proposed MUStARD*}                                        & \textbf{72.1}                                               & \textbf{72}                                     & \textbf{72}                                      & \multirow{-5}{*}{\textit{\textbf{}}} & \textbf{74.2}                                               & \textbf{74.2}                                     & \textbf{74.2}                                      \\ \cline{1-4} \cline{6-8} 

\multicolumn{1}{|l|}{\textbf{$\%\Delta$MUStARD}} & \multicolumn{1}{r}{{\color[HTML]{036400} $\uparrow$ 3.69\%}} & \multicolumn{1}{r}{{\color[HTML]{036400} $\uparrow$ 9.09\%}} & \multicolumn{1}{r|}{{\color[HTML]{036400} $\uparrow$ 9.25\%}} & \multicolumn{1}{l|}{}                & \multicolumn{1}{r}{{\color[HTML]{036400} $\uparrow$ 1.08\%}} & \multicolumn{1}{c}{{\color[HTML]{036400} $\uparrow$ 1.99\%}} & \multicolumn{1}{r|}{{\color[HTML]{036400} $\uparrow$ 2.24\%}} \\ \hline
\textbf{Proposed MUStARD++}                                        & 70.2                                               & 70.2                                     & 70.2                                      & \multirow{-5}{*}{\textit{\textbf{}}} & 70.3                                               & 70.3                                     & 70.3                                      \\ \cline{1-4} \cline{6-8} 
\hline
\end{tabular}%
}
\caption{Sarcasm detection results (weighted average) comparison with SOTA on MUStARD and MUStARD++. Proposed MUStARD is our best model on MUStARD and $\%\Delta$MUStARD is the improvement we observe with our model and corrected labels of MUStARD. Proposed MUStARD++ is results of our best model for sarcasm detection on extended dataset MUStARD++ presented in this paper.}
\label{tab:sarcasm results comparison}
\end{table*}


\begin{table*}[htp!]
\centering
\resizebox{\textwidth}{!}{%
\begin{tabular}{c|ccc|ccc|l|ccc|ccc|}
\cline{2-14}
\multicolumn{1}{l|}{}                & \multicolumn{6}{c|}{\textbf{Speaker Independent}}                                             &  & \multicolumn{6}{c|}{\textbf{Speaker Dependent}}                                               \\ \cline{2-7} \cline{9-14} 
\multicolumn{1}{l|}{}                & \multicolumn{3}{c|}{\textbf{w/o Context}}     & \multicolumn{3}{c|}{\textbf{w Context}}       &  & \multicolumn{3}{c|}{\textbf{w/o Context}}     & \multicolumn{3}{c|}{\textbf{w Context}}       \\ \cline{2-7} \cline{9-14} 
\multicolumn{1}{l|}{}                & \textbf{P}    & \textbf{R}    & \textbf{F1}   & \textbf{P}    & \textbf{R}    & \textbf{F1}   &  & \textbf{P}    & \textbf{R}    & \textbf{F1}   & \textbf{P}    & \textbf{R}    & \textbf{F1}   \\ \cline{1-7} \cline{9-14} 
\multicolumn{1}{|c|}{\textbf{T}}     & 67.9 & 67.7 & 67.7 & 69.3 & 69.2 & 69.2 && 69.4 & 69.3 & 69.3 & 70.2 & 70 & 70          \\
\multicolumn{1}{|c|}{\textbf{A}}     & 63.9          & 63.5          & 63.6          & 64.3          & 64.1          & 64.1          &  & 65.3          & 65.2          & 65.2          & 65.0         & 64.9          & 64.9          \\
\multicolumn{1}{|c|}{\textbf{V}}     &59.5	&59.4	&59.4         & 60.3 &60.0 &60.0          &  & 61.8          & 61.7          & 61.7          & 61.6          &61.4          & 61.5        \\
\multicolumn{1}{|c|}{\textbf{T+A}}   & 68.8 & 68.6 & 68.7 &\textbf{70.2} & \textbf{70.2} & \textbf{70.2} && \textbf{69.8} & \textbf{69.5} & \textbf{69.5} & 69.2 & 69.1 & 69.1                  \\
\multicolumn{1}{|c|}{\textbf{A+V}}   & 65.7& 65.4& 65.5 &67.5 &67.3 &67.4 &&64.9 &64.5 &64.5  &64.2 &64.0 &64.0          \\
\multicolumn{1}{|c|}{\textbf{V+T}}   & 68.2 & 68.1 & 68.1 & 67.9 & 67.6 & 67.6 && 69.1 & 69.0 & 69.0 & 69.4 & 69.1 & 69.1         \\
\multicolumn{1}{|c|}{\textbf{T+A+V}} &\textbf{69.5} & \textbf{69.4} & \textbf{69.4} & 69.6 & 69.5 & 69.6 && 69.6 & 69.3 & 69.3 & \textbf{70.6} & \textbf{70.3} & \textbf{70.3}  \\ \cline{1-7} \cline{9-14} 
\end{tabular}%
}
\captionsetup{justification=centering}
\caption{Sarcasm detection results for \textbf{MUStARD++}, Weighted Average}
\label{tab:Sarcasm detection results for MUStARD++}
\end{table*}


%
\section{Results and Analysis} \label{sec:experiments}
This section discusses the benchmarking experiments done with the proposed dataset for sarcasm detection, emotion recognition and arousal-valence prediction. 

\subsection{Results of Sarcasm Detection}
Table \ref{tab:sarcasm results comparison} 
shows results of our best model for sarcasm detection on both MUStARD and MUStARD++. Our results outperform state-of-art models significantly on MUStARD which demonstrates the superiority of the collaborative gating-based multimodal fusion, and the best modality features selected were BART for text, low-level audio descriptors, and ResNET video features). We performed attention over modalities and intra-modality attention which helped us understand the importance of features. 
Also in \citelanguageresource{ACL2020PB} emotion and sentiment labels are used in the sarcasm detection task, but we are able to outperform them without using emotion or sentiment label. Table \ref{tab:Sarcasm detection results for MUStARD++} shows the best results per modality and modality combination in the speaker independent and speaker dependent setup. The experiments with context and without context clearly show the importance of contextual dialogues to be able to detect sarcasm in an utterance. Although results with the speaker-dependent setting is marginally better in all modality combinations, we believe a speaker-independent setting is better since speaker information might bias the system towards sarcasm. We intend to use this sarcasm detection module as the first module of our system followed by emotion recognition on the sarcastic sentences and valence and arousal prediction to understand the degree of emotions identified, Thus did not use emotion, sentiment, or valence-arousal for sarcasm detection task. However, we plan to use sarcasm detection output and valence-arousal predictions to see if we can improve emotion recognition.

\subsection{Emotion Recognition Results}
Of the various models and features that we used, BART for text, MFCC, spectrogram and prosodic for audio and features learnt from keyframes using ResNET for video worked best for this dataset. 
Due to the small size of this dataset, we pretrained models on IEMOCAP \cite{iemocap} and MELD \cite{MELD} and then tried zero-shot experiments on MUStARD and MUStARD++. Since IEMOCAP and MELD do not have any sarcastic utterances, the models saw a significant drop in F-score when tested on sarcastic data. We also extracted learnt audio features using state-of-art self-supervised PASE network \cite{Pase2019} but models built on PASE features require significantly large sarcastic data although pretrained on different utterances. 

Table \ref{tab:implicit-Emotion distribution in Sarcastic Utterances-multiclass} and Table \ref{tab:explicit-Emotion distribution in Sarcastic Utterances-multiclass} show the detailed emotion classification results on sarcastic utterances for both implicit and explicit emotion in speaker-dependent and speaker-independent setting. Ablation studies across all modalities show that audio and video when used with text perform better. This is intuitive because we consider the variation in speech signals in audio and visual features from video while ignoring the actual spoken content. We observe that for emotion classification, contextual information plays a key role. Although we observe similar numbers in speaker-dependent and speaker-independent setting, it is better to have a speaker-independent setting than limit the overall method by passing speaker's information. 
\begin{table*}[h!]
\centering
\resizebox{\textwidth}{!}{%
\begin{tabular}{c|ccc|ccc|l|ccc|ccc|}
\cline{2-14}
\multicolumn{1}{l|}{}                & \multicolumn{6}{c|}{\textbf{Speaker Independent}}                                             &  & \multicolumn{6}{c|}{\textbf{Speaker Dependent}}                                               \\ \cline{2-7} \cline{9-14} 
\multicolumn{1}{l|}{}                & \multicolumn{3}{c|}{\textbf{w/o Context}}     & \multicolumn{3}{c|}{\textbf{w Context}}       &  & \multicolumn{3}{c|}{\textbf{w/o Context}}     & \multicolumn{3}{c|}{\textbf{w Context}}       \\ \cline{2-7} \cline{9-14} 
\multicolumn{1}{l|}{}                & \textbf{P}    & \textbf{R}    & \textbf{F1}   & \textbf{P}    & \textbf{R}    & \textbf{F1}   &  & \textbf{P}    & \textbf{R}    & \textbf{F1}   & \textbf{P}    & \textbf{R}    & \textbf{F1}   \\ \cline{1-7} \cline{9-14} 
\multicolumn{1}{|c|}{\textbf{T}}     & 33$\pm$0.9          & 33.6$\pm$1          & \textcolor{red}{33.3$\pm$0.9}        & 32.3$\pm$0.7          & 32.7$\pm$0.6         & 32.5$\pm$0.6         &  & 29.9$\pm$0.9          & 30.3$\pm$0.8           & 30.1$\pm$0.8         & 30.2$\pm$0.9          & 30.8$\pm$0.9          & 30.5$\pm$0.9          \\
\multicolumn{1}{|c|}{\textbf{A}}     & 26.7$\pm$1.1          & 27.1$\pm$1.4          & 26.8$\pm$1.1          & 24.9$\pm$1.0          & 26.3$\pm$1.4          & 25.5$\pm$1.2          &  & 24.3$\pm$0.8          & 24.7$\pm$0.6         & 24.5$\pm$0.7          & 26.7$\pm$1.2         & 26.9$\pm$1.2          & 26.8$\pm$1.2          \\
\multicolumn{1}{|c|}{\textbf{V}}     & 28.8$\pm$0.9          & 29.4$\pm$1.3          & 29$\pm$1.1          & 28.5$\pm$1.2          & 29.2$\pm$1.4          & 28.8$\pm$1.3          &  & 30.3$\pm$1.4          & 31.4$\pm$1.2         & 30.6$\pm$1.4          &28.7$\pm$0.8          & 30.08$\pm$1.1          & 29.1$\pm$1.0         \\
\multicolumn{1}{|c|}{\textbf{T+A}}   & 31.5$\pm$1.7          & 31.6$\pm$1.8         & 31.6$\pm$1.7          & 32.1$\pm$0.7        & 32.04$\pm$0.6          & 32.03$\pm$0.6          &  & 29.1$\pm$1.6          &29.2$\pm$1.5          & 29.1$\pm$1.5         & 31.2$\pm$2          & 31.8$\pm$1.8         & 31.4$\pm$1.8         \\
\multicolumn{1}{|c|}{\textbf{A+V}}   & 25.9$\pm$1.9          & 26.3$\pm$2          & 26.1$\pm$1.9         & 28.2$\pm$1.1          & 28.3$\pm$1.2          & 28.2$\pm$1.1          &  & 29.7$\pm$0.6          & 30.6$\pm$0.9          & 30.1$\pm$0.7         & 25.2$\pm$1.0          & 25.2$\pm$0.9          & 25.2$\pm$0.9          \\
\multicolumn{1}{|c|}{\textbf{V+T}}   & 31.9$\pm$0.8   & 32.5$\pm$0.7 & 32.2$\pm$0.7  & 32.7$\pm$1.1  & 33.3$\pm$1.0  & \textcolor{red}{33.0$\pm$1.1} &  & 31.1$\pm$0.8          & 31.2$\pm$0.7         & \textcolor{red}{31.1$\pm$0.7}          & 31.8$\pm$0.         & 31.9$\pm$0.5        & \textcolor{red}{31.8$\pm$0.6}          \\
\multicolumn{1}{|c|}{\textbf{T+A+V}} & 31.2$\pm$1         & 31.6$\pm$0.1        & 31.4$\pm$1         & 28.9$\pm$1.3         & 29.0$\pm$1.4          & 28.9$\pm$1.3          &  & 30.9$\pm$0.3 & 30.5$\pm$0.4 & 30.7$\pm$0.3 & 31.6$\pm$1.5 & 31.3$\pm$1.3 & 31.5$\pm$1.4 \\\cline{1-7} \cline{9-14} 

\hline
\end{tabular}%
}
\captionsetup{justification=centering}
\caption{Mean, std-dev of 5 runs for Implicit Emotion Classification (Multiclass) on MUStARD++}
\label{tab:implicit-Emotion distribution in Sarcastic Utterances-multiclass}
\end{table*}

\begin{table*}[h!]
\centering
\resizebox{\textwidth}{!}{%
\begin{tabular}{c|ccc|ccc|l|ccc|ccc|}
\cline{2-14}
\multicolumn{1}{l|}{}                & \multicolumn{6}{c|}{\textbf{Speaker Independent}}                                             &  & \multicolumn{6}{c|}{\textbf{Speaker Dependent}}                                               \\ \cline{2-7} \cline{9-14} 
\multicolumn{1}{l|}{}                & \multicolumn{3}{c|}{\textbf{w/o Context}}     & \multicolumn{3}{c|}{\textbf{w Context}}       &  & \multicolumn{3}{c|}{\textbf{w/o Context}}     & \multicolumn{3}{c|}{\textbf{w Context}}       \\ \cline{2-7} \cline{9-14} 
\multicolumn{1}{l|}{}                & \textbf{P}    & \textbf{R}    & \textbf{F1}   & \textbf{P}    & \textbf{R}    & \textbf{F1}   &  & \textbf{P}    & \textbf{R}    & \textbf{F1}   & \textbf{P}    & \textbf{R}    & \textbf{F1}   \\ \cline{1-7} \cline{9-14} 
\multicolumn{1}{|c|}{\textbf{T}}     & 38.5$\pm$0.8          & 39.2$\pm$0.9          & 38.8$\pm$0.8          & 38.2$\pm$1.2          & 38.8$\pm$1.3          & 38.5$\pm$1.3         &  & 38.7$\pm$0.5          & 39.3$\pm$0.5          & 39$\pm$0.5          & 38.9$\pm$0.4          & 39.7$\pm$0.6          & 39.3$\pm$0.5          \\
\multicolumn{1}{|c|}{\textbf{A}}     & 26.4$\pm$0.9          & 28$\pm$1.4        & 27.1$\pm$1  & 27.1$\pm$1.1         & 28.1$\pm$1.3         & 27.6$\pm$1.2          && 28.1$\pm$1.1          &  29.3$\pm$1.0          & 28.6$\pm$1.1         & 28.4$\pm$0.7          & 31.6$\pm$1.3          & 29.6$\pm$0.8                  \\
\multicolumn{1}{|c|}{\textbf{V}}     & 25.1$\pm$0.6          & 25.9$\pm$0.6         & 25.5$\pm$0.6        & 24.4$\pm$0.7          & 24.9$\pm$0.9         & 24.6$\pm$0.8          &  & 25.7$\pm$1.4          & 36.1$\pm$0.8          & 27.7$\pm$0.8          & 27.0$\pm$0.8          & 29.6$\pm$1.6          & 28.0$\pm$1         \\
\multicolumn{1}{|c|}{\textbf{T+A}}   & 38.9$\pm$1.1          & 39.5$\pm$1.3          & \textcolor{red}{39.2$\pm$1.2}          & 39.2$\pm$0.6          & 39.5$\pm$0.6          & 39.3$\pm$0.6          &  & 39.1$\pm$0.8          & 39.7$\pm$0.7          & 39.4$\pm$0.7          & 39.1$\pm$0.5          & 39.5$\pm$0.7          & 39.3$\pm$0.6          \\
\multicolumn{1}{|c|}{\textbf{A+V}}   & 26.4$\pm$1.4          & 26.5$\pm$1.5          & 26.4$\pm$1.4          & 26.2$\pm$1.22          & 26.3$\pm$1.6          & 26.2$\pm$1.5          &          & 27.6$\pm$1          & 28.2$\pm$1.2          & 27.9$\pm$1.1          & 27.8$\pm$0.5          & 28$\pm$0.4 &      27.9$\pm$0.4     \\
\multicolumn{1}{|c|}{\textbf{V+T}}   & 38.6$\pm$0.7 & 39.2$\pm$0.8 & 38.8$\pm$0.8 & 40.5$\pm$0.7 & 41.2$\pm$0.7 & \textcolor{red}{40.8$\pm$0.7} &  & 39.8$\pm$0.1 & 40$\pm$0.2 & 39.8$\pm$0.2 & 39.9$\pm$0.4 & 40.3$\pm$0.6 & \textcolor{red}{40$\pm$0.5}  \\
\multicolumn{1}{|c|}{\textbf{T+A+V}} & 37.8$\pm$0.1          & 38.3$\pm$0.8          & 38.0$\pm$0.9          & 39.5$\pm$0.8          & 39.6$\pm$0.9          &  39.5$\pm$0.9 &  & 40$\pm$0.6  & 39.8$\pm$0.5&   \textcolor{red}{39.9$\pm$0.9}           & 39.7$\pm$1.3         & 39.4$\pm$1.2          & 39.5$\pm$1.2         \\ \cline{1-7} \cline{9-14} 
\hline
\end{tabular}%
}
\captionsetup{justification=centering}
\caption{Mean, std-dev of 5 runs for Explicit Emotion Classification (Multiclass) on MUStARD++}
\label{tab:explicit-Emotion distribution in Sarcastic Utterances-multiclass}
\end{table*}


\begin{table*}[ht!]
\resizebox{\textwidth}{!}{%
\begin{tabular}{c|cc|cc|cc|cc|l|cc|cc|cc|cc|}
\cline{2-18}
\multicolumn{1}{l|}{}                     & \multicolumn{8}{c|}{\textbf{Valence Prediction}}                                                                                                                          &  & \multicolumn{8}{c|}{\textbf{Arousal Prediction}}                                                                                                                          \\ \cline{2-9} \cline{11-18} 
\multicolumn{1}{l|}{\textbf{}}            & \multicolumn{4}{c|}{\textbf{Speaker Independent}}                                   & \multicolumn{4}{c|}{\textbf{Speaker Dependent}}                                     &  & \multicolumn{4}{c|}{\textbf{Speaker Independent}}                                   & \multicolumn{4}{c|}{\textbf{Speaker Dependent}}                                     \\ \cline{2-9} \cline{11-18} 
\multicolumn{1}{l|}{}                     & \multicolumn{2}{c|}{\textbf{w/o Context}} & \multicolumn{2}{c|}{\textbf{w Context}} & \multicolumn{2}{c|}{\textbf{w/o Context}} & \multicolumn{2}{c|}{\textbf{w Context}} &  & \multicolumn{2}{c|}{\textbf{w/o Context}} & \multicolumn{2}{c|}{\textbf{w Context}} & \multicolumn{2}{c|}{\textbf{w/o Context}} & \multicolumn{2}{c|}{\textbf{w Context}} \\ \cline{2-9} \cline{11-18} 
\multicolumn{1}{l|}{}                     & \textbf{MAE}        & \textbf{RMSE}        & \textbf{MAE}       & \textbf{RMSE}       & \textbf{MAE}        & \textbf{RMSE}        & \textbf{MAE}       & \textbf{RMSE}       &  & \textbf{MAE}        & \textbf{RMSE}        & \textbf{MAE}       & \textbf{RMSE}       & \textbf{MAE}        & \textbf{RMSE}        & \textbf{MAE}       & \textbf{RMSE}       \\ \cline{1-9} \cline{11-18} 
\multicolumn{1}{|c|}{{\textbf{T}}}  &0.91 &0.74 &0.89 &0.73 &0.98 &0.78
&0.95 &0.75 &&  1.42 &1.15 &1.45 &1.16 &1.42 &1.10 &1.42 &1.12
              \\
\multicolumn{1}{|c|}{\textbf{A}} & 0.82 &0.65 &0.91 &0.73 &0.97 &0.77 &0.95 &0.76 &&1.24 &1.00 &1.34 &1.07 &1.45 &1.15 &1.43 &1.11
              \\
\multicolumn{1}{|c|}{\textbf{V}} &0.85 &0.68 &0.84 &0.68 &0.94 &0.76 &1.02
&0.81 &&1.22 &0.96 &1.16 &0.93 &1.45 &1.14 &1.47 &1.16
\\
\multicolumn{1}{|c|}{\textbf{T+A}}  &0.83 &0.67 &0.86 &0.69 &0.93 &0.74 &0.91 &0.73 &&1.18 &0.96 &1.33 &1.08 &1.41 &1.11 &1.59 &1.23
             \\
\multicolumn{1}{|c|}{\textbf{A+V}}  &0.83 &0.66 &0.82 &0.66 &0.92 &0.79 &0.88 &0.71 &&1.19 &0.96 &1.21 &0.98 &1.37 &1.11 &1.43 &1.16
              \\
\multicolumn{1}{|c|}{\textbf{V+T}} &0.82 &0.66 &0.86 &0.71 &0.90 &0.72 &0.93 &0.74 &&1.22 &0.97 &1.34 &1.09 &1.50 &1.12 &1.56 &1.24
              \\
\multicolumn{1}{|c|}{\textbf{T+A+V}}  &0.78 &0.62 &0.78 &0.62 &0.85 &0.70 &0.94 &0.75 &&1.20 &0.95 &1.20 &0.95 &1.23 &0.98 &1.45 &1.10
 \\ \hline
\end{tabular}%
}
\caption{Valence-Arousal Prediction on MUStARD++ (Mean of 5 runs)}
\label{tab:Resultsval-aro}
\end{table*}


Post hyper-parameter search, best parameters in a 5-fold cross validation is selected across 28 different experimental configurations: 7 modality combinations (rows of Table \ref{tab:implicit-Emotion distribution in Sarcastic Utterances-multiclass}) each run with 4 settings (columns of Table \ref{tab:implicit-Emotion distribution in Sarcastic Utterances-multiclass}).
Table \ref{tab:Resultsval-aro} shows results of valence and arousal predictions across different modalities Here also we observed that in a speaker-independent setting the model performs better, but as the length of the speaker vector increases with more people, the effect of speaker information confuses the models. We also observe that contextual information doesn't affect valence arousal prediction as that of the actual utterance, which is intuitive.

In order to prove the performance improvement due to correcting labels, we ran a Wilcoxon signed rank test \cite{Wilcoxon} on old versus new labeled data. We made 10 runs with the best trained multimodal model on old as well as new labels on the sarcastic MUStARD data for proper comparison. The mean F1 score of emotion recognition on the sarcastic set with old labels is $20.64 \pm 1.15$ while with new labels on same set the mean is $39.5 \pm 1.00$ which is statistically significant with a p-value of 0.002. We observed reduction in label confusion in the confusion matrix. Thus these re-labeling efforts added trust to the extended MUStARD++ dataset annotated using same label set.

\section{Conclusions and Future Work} \label{sec:conslusion}
This paper presents a multimodal sarcasm dataset that can be used by researchers in the area of sarcasm detection and emotion recognition. We start with the version of MUStARD data provided by \citelanguageresource{ACL2020PB} and correct several emotion labels, while appending the sarcasm type and arousal-valence labels as additional metadata that is useful for both the research avenues. We doubled the number of sarcastic videos, finding which is very challenging, thus making it a beneficial contribution to the research community. We also added equal number of non-sarcastic videos with their context along with similar metadata annotations. To the best of our knowledge this is the first work on emotion recognition in sarcasm and towards that we present a curated dataset which is benchmarked using several pretrained feature extractors and multimodal fusion techniques in different setups. 
Sarcasm type information enables a multimodal system to choose the right modality combination for a given utterance, thereby optimizing performance of the sarcasm detector and the emotion recognizer for different utterances. In this work, we have used arousal only to understand the degree/intensity of an emotion, for example, annoyance to anger to rage. In future we want to use arousal and valence to investigate its effect on emotion classification.
While we currently explore the use of sarcasm label in emotion recognition, an interesting research direction would be using emotion labels to improve sarcasm detection. 

\section{Ethical Considerations in Data Curation}
This paper does not claim to find the exact intended emotion of the speaker that led to a sarcastic sentence. Rather we try to predict the perceived emotion. Our annotators annotate on the recorded video, thus observing the perceived emotion, arousal and valence. This is very important for conversational systems where the bot needs to understand the emotion, valence and intensity to be able to respond better. Also, this is in accordance to the suggestions in the ethics sheet for automated emotion recognition \cite{EthicsSheet} where authors explain that given the complexity of human emotion, it is very difficult to predict the exact emotional state of the speaker.  
The authors hereby acknowledge that there could be a possibility of bias in the final emotion label assigned since the label is chosen based on majority voting. In order to minimize the effect of bias, we collect videos from a diverse set of sources and ask seven annotators of different age, gender, and educational background to label their perceived emotions. We have considered the guidelines \cite{EthicsSheet} for responsible development and use of Automated Emotion Recognition systems (AER) and adhered to them in our research statement, data collection, annotation protocol, and during the benchmarking experiments.

\section*{References}\label{reference}

\bibliographystyle{lrec2022-bib}
\bibliography{lrec2022-example}

\begin{thebibliography}{}

\bibitem[\protect\citename{Castro \bgroup et al.\egroup }2019]{mustard}
Castro, Santiago and Hazarika, Devamanyu and P{\'e}rez-Rosas, Ver{\'o}nica and
  Zimmermann, Roger and Mihalcea, Rada and Poria, Soujanya.
\newblock (2019).
\newblock {\em Towards Multimodal Sarcasm Detection (An \_Obviously\_ Perfect
  Paper)}.

\bibitem[\protect\citename{Chauhan \bgroup et al.\egroup }2020]{ACL2020PB}
Chauhan, Dushyant Singh and Dhanush, SR and Ekbal, Asif and Bhattacharyya,
  Pushpak.
\newblock (2020).
\newblock {\em Sentiment and Emotion help Sarcasm? A Multi-task Learning
  Framework for Multi-Modal Sarcasm, Sentiment and Emotion Analysis}.

\end{thebibliography}


\begin{thebibliography}{}

\bibitem[\protect\citename{Bedi \bgroup et al.\egroup }2021]{msh}
Bedi, M., Kumar, S., Akhtar, M.~S., and Chakraborty, T.
\newblock (2021).
\newblock Multi-modal sarcasm detection and humor classification in code-mixed
  conversations.
\newblock {\em CoRR}, abs/2105.09984.

\bibitem[\protect\citename{Busso \bgroup et al.\egroup }2008]{iemocap}
Busso, C., Bulut, M., Lee, C.-C., Kazemzadeh, A., Mower, E., Kim, S., Chang,
  J.~N., Lee, S., and Narayanan, S.~S.
\newblock (2008).
\newblock Iemocap: Interactive emotional dyadic motion capture database.
\newblock {\em Language resources and evaluation}, 42(4):335.

\bibitem[\protect\citename{Cai \bgroup et al.\egroup
  }2019]{cai-etal-2019-multi}
Cai, Y., Cai, H., and Wan, X.
\newblock (2019).
\newblock Multi-modal sarcasm detection in {T}witter with hierarchical fusion
  model.
\newblock In {\em Proceedings of the 57th Annual Meeting of the Association for
  Computational Linguistics}, pages 2506--2515, Florence, Italy, July.
  Association for Computational Linguistics.

\bibitem[\protect\citename{Camp}2012]{camp2012sarcasm}
Camp, E.
\newblock (2012).
\newblock Sarcasm, pretense, and the semantics/pragmatics distinction.
\newblock {\em No{\^u}s}, 46(4):587--634.

\bibitem[\protect\citename{Chen \bgroup et al.\egroup }2018]{EmotionLines}
Chen, S., Hsu, C., Kuo, C., Huang, T.~K., and Ku, L.
\newblock (2018).
\newblock Emotionlines: An emotion corpus of multi-party conversations.
\newblock {\em CoRR}, abs/1802.08379.

\bibitem[\protect\citename{Cowie and Cornelius}2003]{cowie2003describing}
Cowie, R. and Cornelius, R.~R.
\newblock (2003).
\newblock Describing the emotional states that are expressed in speech.
\newblock {\em Speech communication}, 40(1-2):5--32.

\bibitem[\protect\citename{Devlin \bgroup et al.\egroup }2018]{bert}
Devlin, J., Chang, M.-W., Lee, K., and Toutanova, K.
\newblock (2018).
\newblock Bert: Pre-training of deep bidirectional transformers for language
  understanding.
\newblock {\em arXiv preprint arXiv:1810.04805}.

\bibitem[\protect\citename{Ekman}1999]{ekman}
Ekman, P.
\newblock (1999).
\newblock Basic emotions.
\newblock {\em Handbook of cognition and emotion}, 98(45-60):16.

\bibitem[\protect\citename{{He} \bgroup et al.\egroup }2016]{ResNet}
{He}, K., {Zhang}, X., {Ren}, S., and {Sun}, J.
\newblock (2016).
\newblock Deep residual learning for image recognition.
\newblock In {\em 2016 IEEE Conference on Computer Vision and Pattern
  Recognition (CVPR)}.

\bibitem[\protect\citename{Jadon and Jasim}2019]{jadon2019video}
Jadon, S. and Jasim, M.
\newblock (2019).
\newblock Video summarization using keyframe extraction and video skimming.
\newblock {\em arXiv preprint arXiv:1910.04792}.

\bibitem[\protect\citename{Joshi \bgroup et al.\egroup
  }2016]{DBLP:journals/corr/JoshiBC16}
Joshi, A., Bhattacharyya, P., and Carman, M.~J.
\newblock (2016).
\newblock Automatic sarcasm detection: {A} survey.
\newblock {\em CoRR}, abs/1602.03426.

\bibitem[\protect\citename{Joshi \bgroup et al.\egroup }2017]{surveyJoshi}
Joshi, A., Bhattacharyya, P., and Carman, M.~J.
\newblock (2017).
\newblock Automatic sarcasm detection: A survey.
\newblock {\em ACM Comput. Surv.}, 50(5), September.

\bibitem[\protect\citename{Joshi \bgroup et al.\egroup
  }2018]{SarcasmReviewBook}
Joshi, A., Bhattacharyya, P., and Carman, M.~J.
\newblock (2018).
\newblock {\em Investigations in Computational Sarcasm}.

\bibitem[\protect\citename{Khodak \bgroup et al.\egroup
  }2017]{DBLP:journals/corr/KhodakSV17}
Khodak, M., Saunshi, N., and Vodrahalli, K.
\newblock (2017).
\newblock A large self-annotated corpus for sarcasm.
\newblock {\em CoRR}, abs/1704.05579.

\bibitem[\protect\citename{Lewis \bgroup et al.\egroup }2019]{BART}
Lewis, M., Liu, Y., Goyal, N., Ghazvininejad, M., Mohamed, A., Levy, O.,
  Stoyanov, V., and Zettlemoyer, L.
\newblock (2019).
\newblock {BART:} denoising sequence-to-sequence pre-training for natural
  language generation, translation, and comprehension.
\newblock {\em CoRR}, abs/1910.13461.

\bibitem[\protect\citename{Liu \bgroup et al.\egroup }2019]{RoBERTa}
Liu, Y., Ott, M., Goyal, N., Du, J., Joshi, M., Chen, D., Levy, O., Lewis, M.,
  Zettlemoyer, L., and Stoyanov, V.
\newblock (2019).
\newblock Roberta: {A} robustly optimized {BERT} pretraining approach.
\newblock {\em CoRR}, abs/1907.11692.

\bibitem[\protect\citename{Liu \bgroup et al.\egroup }2020]{liu2020use}
Liu, Y., Albanie, S., Nagrani, A., and Zisserman, A.
\newblock (2020).
\newblock Use what you have: Video retrieval using representations from
  collaborative experts.

\bibitem[\protect\citename{Maynard and Greenwood}2014]{maynard2014cares}
Maynard, D.~G. and Greenwood, M.~A.
\newblock (2014).
\newblock Who cares about sarcastic tweets? investigating the impact of sarcasm
  on sentiment analysis.
\newblock In {\em LREC 2014 Proceedings}.

\bibitem[\protect\citename{McFee \bgroup et al.\egroup
  }2022]{brian_mcfee_2022_6097378}
McFee, B., Metsai, A., McVicar, M., Balke, S., Thomé, C., Raffel, C., Zalkow,
  F., Malek, A., Dana, Lee, K., Nieto, O., Ellis, D., Mason, J., Battenberg,
  E., Seyfarth, S., Yamamoto, R., viktorandreevichmorozov, Choi, K., Moore, J.,
  Bittner, R., Hidaka, S., Wei, Z., nullmightybofo, Weiss, A., Hereñú, D.,
  Stöter, F.-R., Friesch, P., Vollrath, M., Kim, T., and Thassilo.
\newblock (2022).
\newblock librosa/librosa: 0.9.1, February.

\bibitem[\protect\citename{Mohammad}2021]{EthicsSheet}
Mohammad, S.~M.
\newblock (2021).
\newblock Ethics sheet for automatic emotion recognition and sentiment
  analysis.
\newblock {\em CoRR}, abs/2109.08256.

\bibitem[\protect\citename{Oprea and Magdy}2020]{oprea-magdy-2020-isarcasm}
Oprea, S. and Magdy, W.
\newblock (2020).
\newblock i{S}arcasm: A dataset of intended sarcasm.
\newblock In {\em Proceedings of the 58th Annual Meeting of the Association for
  Computational Linguistics}, pages 1279--1289.

\bibitem[\protect\citename{Oraby \bgroup et al.\egroup
  }2016]{oraby-etal-2016-creating}
Oraby, S., Harrison, V., Reed, L., Hernandez, E., Riloff, E., and Walker, M.
\newblock (2016).
\newblock Creating and characterizing a diverse corpus of sarcasm in dialogue.
\newblock In {\em Proceedings of the 17th Annual Meeting of the Special
  Interest Group on Discourse and Dialogue}, pages 31--41, Los Angeles,
  September. Association for Computational Linguistics.

\bibitem[\protect\citename{Pascual \bgroup et al.\egroup }2019]{Pase2019}
Pascual, S., Ravanelli, M., Serrà, J., Bonafonte, A., and Bengio, Y.
\newblock (2019).
\newblock {Learning Problem-Agnostic Speech Representations from Multiple
  Self-Supervised Tasks}.
\newblock In {\em INTERSPEECH, 2019}, pages 161--165.

\bibitem[\protect\citename{Plutchick}1980]{plutchick-emotion}
Plutchick, R.
\newblock (1980).
\newblock Emotion: a psychoevolutionary synthesis.
\newblock {\em New York, Happer \& Row}.

\bibitem[\protect\citename{Poria \bgroup et al.\egroup }2018]{MELD}
Poria, S., Hazarika, D., Majumder, N., Naik, G., Cambria, E., and Mihalcea, R.
\newblock (2018).
\newblock {MELD:} {A} multimodal multi-party dataset for emotion recognition in
  conversations.
\newblock {\em CoRR}, abs/1810.02508.

\bibitem[\protect\citename{Preo{\c{t}}iuc-Pietro \bgroup et al.\egroup
  }2016]{fb-va}
Preo{\c{t}}iuc-Pietro, D., Schwartz, H.~A., Park, G., Eichstaedt, J., Kern, M.,
  Ungar, L., and Shulman, E.
\newblock (2016).
\newblock Modelling valence and arousal in facebook posts.
\newblock In {\em Proceedings of the 7th workshop on computational approaches
  to subjectivity, sentiment and social media analysis}, pages 9--15.

\bibitem[\protect\citename{Raffel \bgroup et al.\egroup }2020]{T5}
Raffel, C., Shazeer, N.~M., Roberts, A., Lee, K., Narang, S., Matena, M., Zhou,
  Y., Li, W., and Liu, P.~J.
\newblock (2020).
\newblock Exploring the limits of transfer learning with a unified text-to-text
  transformer.
\newblock {\em ArXiv}, abs/1910.10683.

\bibitem[\protect\citename{Russell}1980]{russell-emotion}
Russell, J.
\newblock (1980).
\newblock A circumplex model of affect.
\newblock {\em Journal of Personality and Social Psychology}, 39:1161--1178,
  12.

\bibitem[\protect\citename{Sangwan \bgroup et al.\egroup
  }2020]{multimodal_image}
Sangwan, S., Akhtar, M.~S., Behera, P., and Ekbal, A.
\newblock (2020).
\newblock I didn’t mean what i wrote! exploring multimodality for sarcasm
  detection.
\newblock In {\em 2020 International Joint Conference on Neural Networks
  (IJCNN)}, pages 1--8.

\bibitem[\protect\citename{Schifanella \bgroup et al.\egroup
  }2016]{Schifanella_2016}
Schifanella, R., de~Juan, P., Tetreault, J., and Cao, L.
\newblock (2016).
\newblock Detecting sarcasm in multimodal social platforms.
\newblock {\em Proceedings of the 24th ACM international conference on
  Multimedia}, Oct.

\bibitem[\protect\citename{Wilcoxon}1945]{Wilcoxon}
Wilcoxon, F.
\newblock (1945).
\newblock Individual comparisons by ranking methods.
\newblock {\em Biometrics Bulletin}.

\bibitem[\protect\citename{Zafeiriou \bgroup et al.\egroup }2017]{aff-wild}
Zafeiriou, S., Kollias, D., Nicolaou, M., Papaioannou, A., Zhao, G., and
  Kotsia, I.
\newblock (2017).
\newblock Aff-wild: Valence and arousal ‘in-the-wild’ challenge.
\newblock pages 1980--1987, 07.

\bibitem[\protect\citename{Zahiri and Choi}2017]{emoryNLP}
Zahiri, S.~M. and Choi, J.~D.
\newblock (2017).
\newblock Emotion detection on {TV} show transcripts with sequence-based
  convolutional neural networks.
\newblock {\em CoRR}, abs/1708.04299.

\bibitem[\protect\citename{Zvolenszky}2012]{zvolenszky2012gricean}
Zvolenszky, Z.
\newblock (2012).
\newblock A gricean rearrangement of epithets.

\end{thebibliography}

\section*{Language Resource References}
\bibliographystylelanguageresource{lrec2022-bib}
\bibliographylanguageresource{languageresource}
\section*{Supplementary Material} \label{'sec:appendix'}

\begin{figure*}[!h]
\centering
\includegraphics[width=.3\textwidth]{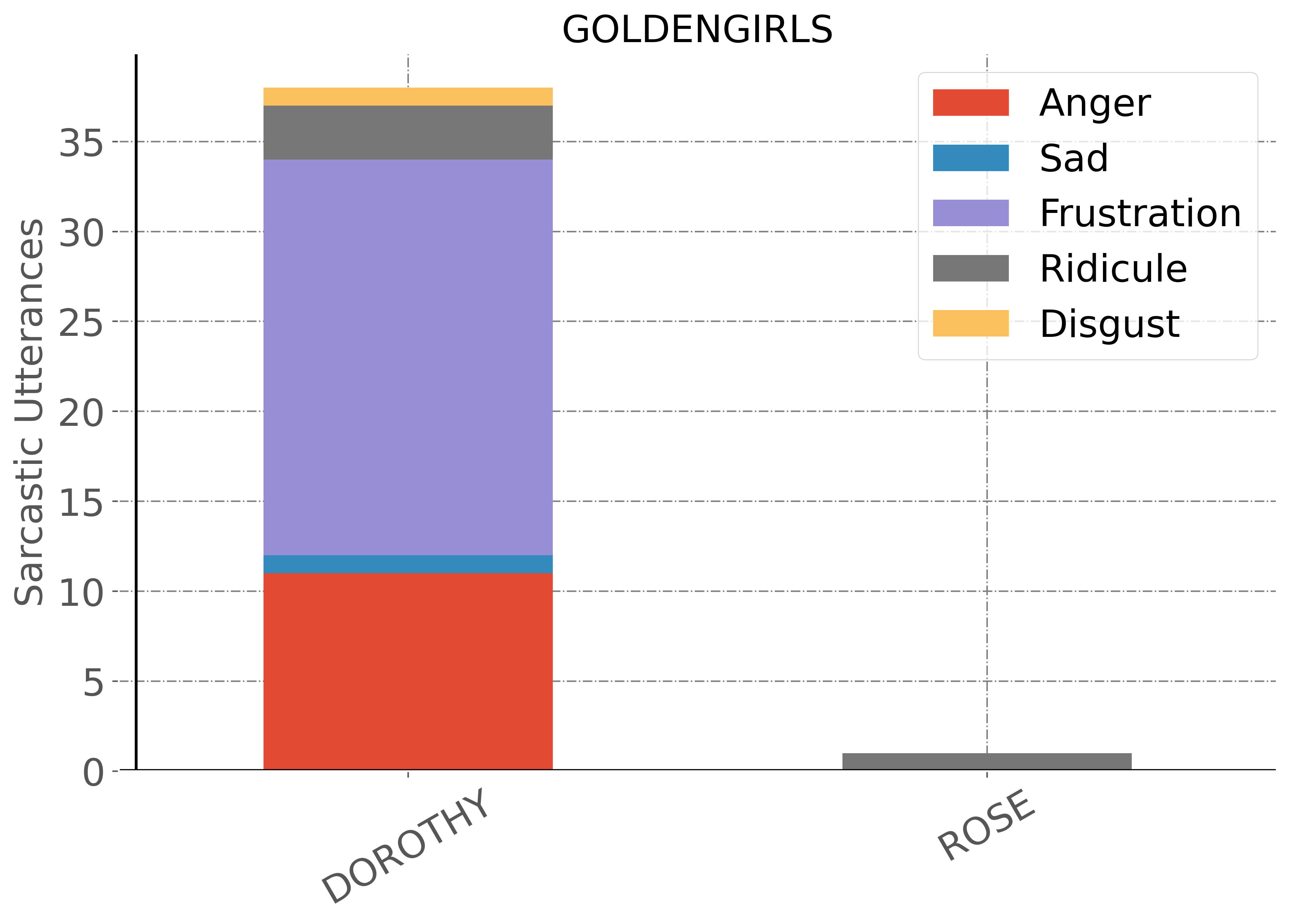}\quad
\includegraphics[width=.3\textwidth]{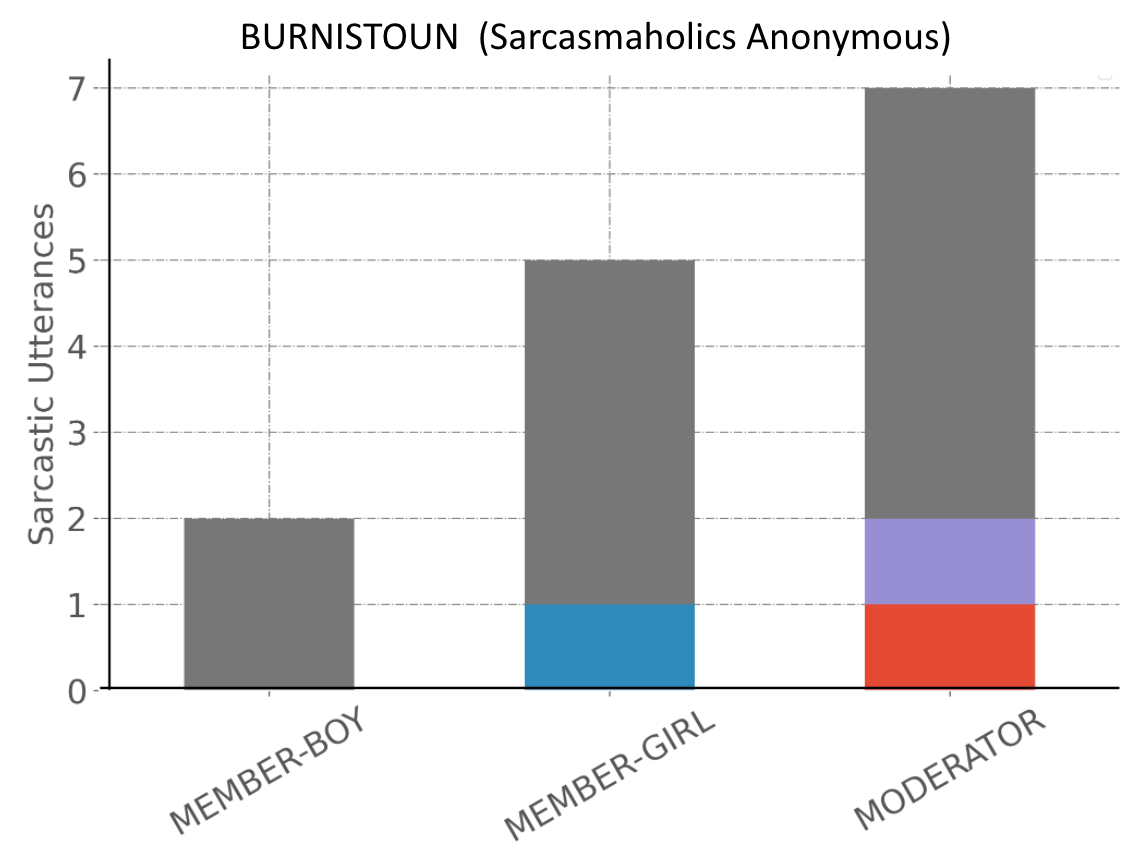}\quad
\includegraphics[width=.3\textwidth]{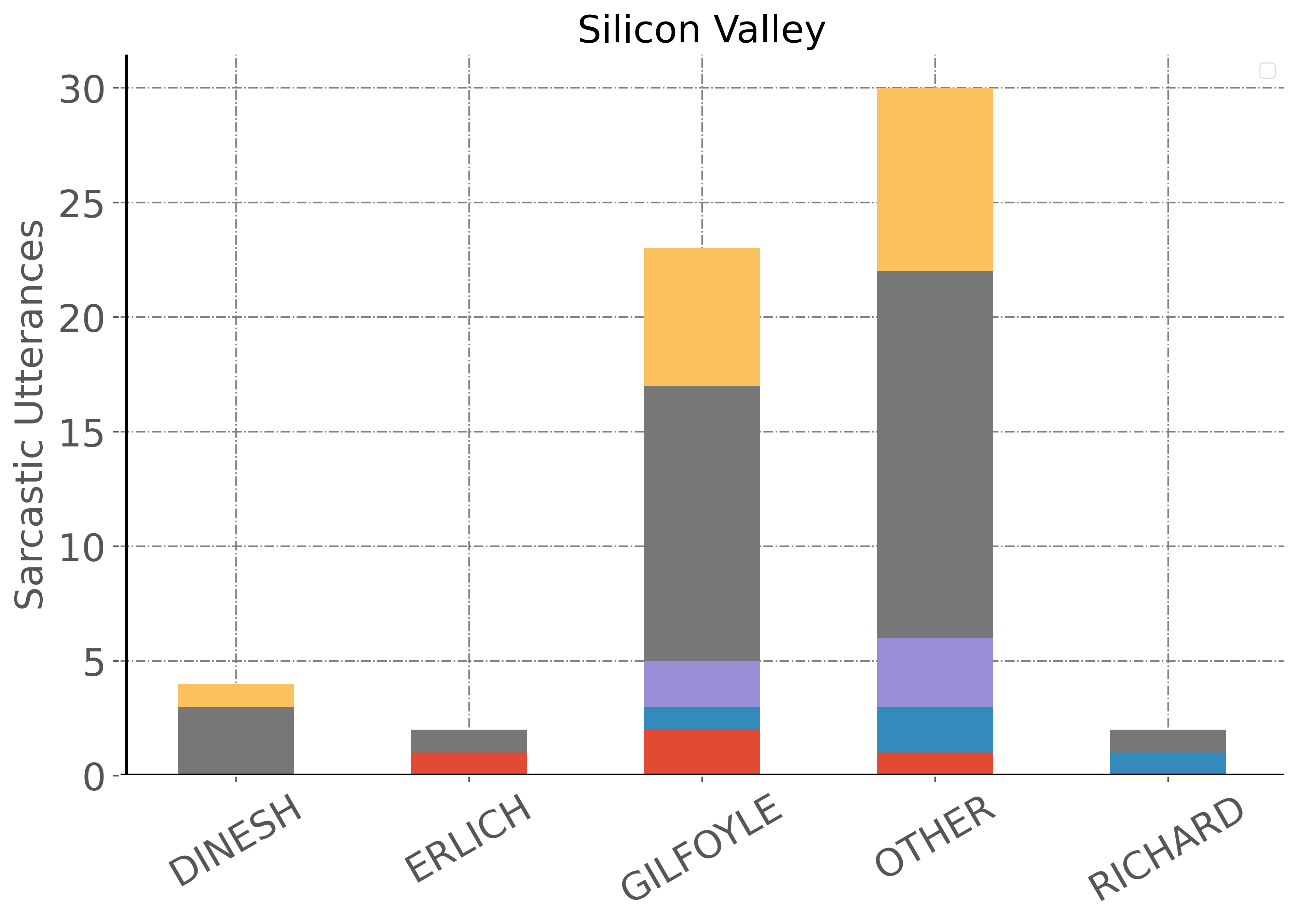}

\medskip

\includegraphics[width=.3\textwidth]{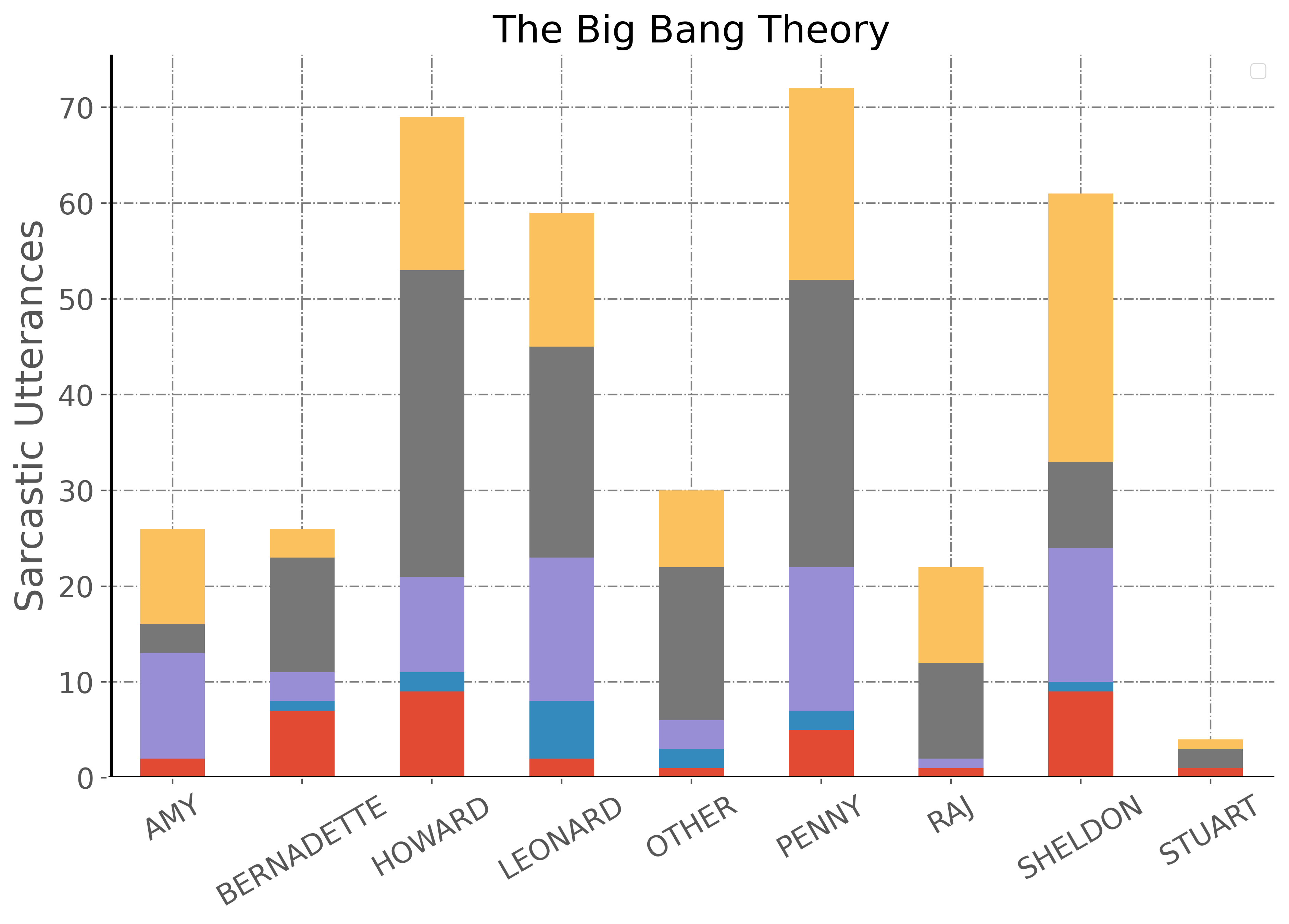}\quad
\includegraphics[width=.3\textwidth]{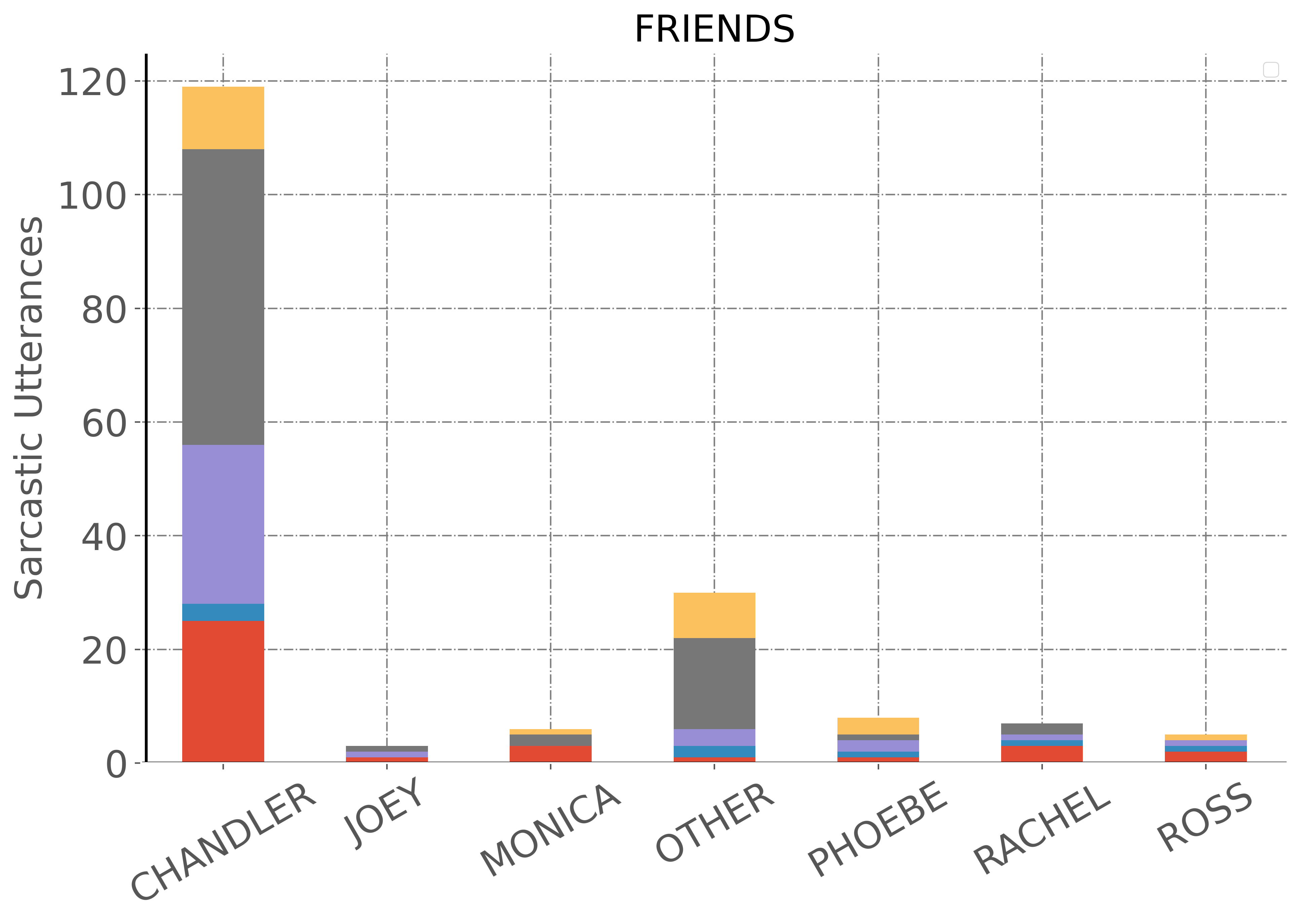}

\caption{Character-label ratio per source  over Implicit Emotion in sarcastic utterances  on MUStARD++ data}
\label{fig:clr}
\end{figure*}

This supplementary material is intended to provide additional information with regards to characteristics of our data, additional analysis and detailed results of experiments.

\subsection*{Additional Data Statistics and Analysis }\label{'sec:additional_analysis'}

Table \ref{tab:emo-dis-complete} shows the distribution of implicit and explicit emotions in our dataset including sarcastic and non-sarcastic utterances.

\begin{table}[ht!]
\resizebox{\columnwidth}{!}{%
\begin{tabular}{|l|c|c|c|c|c|c|c|c|c|c|}
\hline
                 & \textbf{An} & \textbf{Ex} & \textbf{Fe} & \textbf{Sa} & \textbf{Sp} & \textbf{Fs} & \textbf{Hp} & \textbf{Neu} & \textbf{Dis} & \textbf{Ri} \\ \hline
\textbf{Explicit-NS} & 49                    & 47                   & 22                    & 82                   & 28                   & 44                    & 109                  & 190                   & 30                     & 0                    \\
\textbf{Explicit-S} & 4                    & 68                   & 1                    & 67                   & 73                   & 4                    & 135                  & 248                   & 1                     & 0                    \\
\textbf{Implicit-NS} & 50                   & 50                    & 25                    & 89                   & 30                    & 46                  & 111                    & 167                     & 33                   & 0                  \\
\textbf{Implicit-S} & 87                   & 0                    & 0                    & 24                   & 0                    & 130                  & 0                    & 0                     & 134                   & 226                  \\\hline
\end{tabular}%
}
\caption{Emotion distribution in MUStARD++. \footnotesize \{NS-Non sarcastic, S-Sarcastic, An-Anger, Ex- Excitement, Fe-Fear, Sa-Sad, Sp-Surprise, Fs-Frustrated, Hp-Happy Neu-Neutral, Dis-Disgust, Ri-Ridicule\}}
\label{tab:emo-dis-complete}
\end{table}

From the valence-arousal ratings of the dataset, it was observed that all sarcastic utterances received lower valence values, inclined towards the unpleasant end of the spectrum. Non-sarcastic utterances, however, have a more diverse set of valence values with ratings from both pleasant and unpleasant halves of the spectrum. While arousal values in sarcastic utterances have majority ratings as 7-8, in non-sarcastic utterances the distribution is more diverse in comparison with sarcastic utterances and they have majority ratings fall into the range of 5-7. This shows the implied negativity and intensity that sarcasm usually tends to portray. 

In each of the 5 TV shows which form the source of our dataset, different characters contribute to the sarcastic instances with different emotions. To understand the number of speakers in each show and their contribution to various implicit emotions in the sarcastic subset of our data, Figure \ref{fig:clr} provides the distribution of implicit emotions in the utterances of each character from the shows.

For sarcastic utterances, the dataset contains sarcasm type metadata which can help provide deeper insights into the task of sarcasm detection. To also understand how these sarcasm types co-occur with different implicit emotions, Table \ref{tab:emotion in sar-type} presents the distribution of  sarcastic utterances belonging to different implicit emotion classes fall under different sarcasm types.
\begin{table}[ht!]
\centering
\begin{tabular}{|r|c|c|c|c|c|c|}
\hline
\multicolumn{1}{|c|}{} & \multicolumn{1}{l|}{\textbf{An}} & \multicolumn{1}{l|}{\textbf{Sa}} & \textbf{Fr} & \textbf{Ri} & \textbf{Di} &\cellcolor{lightgray} \textbf{Total} \\ \cline{1-7} 
\textbf{PRO} & 47 & 15 & 61 & 129 & 81 & \cellcolor{lightgray}333 \\
\textbf{ILL} & 26 & 4  & 48 & 70  & 30 & \cellcolor{lightgray}178 \\
\textbf{EMB} & 13 & 5  & 20 & 27  & 22 &\cellcolor{lightgray} 87 \\
\textbf{LIK} & 1  & 0  & 1  & 0   & 1 & \cellcolor{lightgray}3  \\ \hline
\end{tabular}%
\caption{Implicit Emotion Distribution per Sarcasm Type. \footnotesize \{An-Anger, Sa-Sad, Fs-Frustrated, Dis-Disgust, Ri-Ridicule\} and \{PRO-Propositional, ILL-Illocutionary, EMB-Embedded, LIK-Likeprefixed\}}
\label{tab:emotion in sar-type}
\end{table}

\begin{table}[h!]
\resizebox{\columnwidth}{!}{%
\begin{tabular}{r|ccc|c|ccc|}
\cline{2-8}
\multicolumn{1}{c|}{\multirow{2}{*}{}}          & \multicolumn{3}{c|}{\textit{\textbf{Implicit}}}                                 & \multirow{7}{*}{\textit{\textbf{}}} & \multicolumn{3}{c|}{\textit{\textbf{Explicit}}}                                 \\ \cline{2-4} \cline{6-8} 
\multicolumn{1}{c|}{}                           & \textbf{P}               & \textbf{R}               & \textbf{F1}               &                                     & \textbf{P}               & \textbf{R}               & \textbf{F1}               \\ \cline{1-4} \cline{6-8} 
\multicolumn{1}{|r|}{\textbf{Majority}}         & 14.1                     & 37.6                     & 20.6                      &                                     & 17.0                     & 41.3                     & 24.1                      \\
\multicolumn{1}{|r|}{\textbf{Random (Uniform)}} & 25.3                     & 20.3                     & 21.4                      &                                     & 27.9                     & 11.8                     & 15.5                      \\
\multicolumn{1}{|r|}{\textbf{Random (Prior)}}   & 25.9                     & 25.9                     & 25.9                      &                                     & 26.1                     & 26.1                     & 26.1                      \\
\multicolumn{1}{|r|}{\textbf{ONEvsREST}}        & \multicolumn{1}{l}{35.6} & \multicolumn{1}{l}{37.6} & \multicolumn{1}{l|}{36.5} &                                     & \multicolumn{1}{l}{41.2
}   & \multicolumn{1}{l}{	43.6}   & \multicolumn{1}{l|}{	42.0}   \\     
\multicolumn{1}{|r|}{\textbf{MultiClass}}       & \multicolumn{1}{r}{34.4} & \multicolumn{1}{r}{36.1} & \multicolumn{1}{r|}{34.9} &                                     & \multicolumn{1}{r}{41.7} & \multicolumn{1}{r}{44.0} & \multicolumn{1}{r|}{42.6} \\
\hline
\end{tabular}%
    }
    \captionsetup{justification=centering}
    \caption{Benchmarking emotion classification results comparison across different methods}
    \label{tab:result-comparision}
\end{table}

\begin{table*}[ht!]
\centering
\resizebox{\textwidth}{!}{%
\begin{tabular}{c|ccc|ccc|l|ccc|ccc|}
\cline{2-14}
\multicolumn{1}{l|}{}                & \multicolumn{6}{c|}{\textbf{Speaker Independent}}                                             &  & \multicolumn{6}{c|}{\textbf{Speaker Dependent}}                                               \\ \cline{2-7} \cline{9-14} 
\multicolumn{1}{l|}{}                & \multicolumn{3}{c|}{\textbf{w/o Context}}     & \multicolumn{3}{c|}{\textbf{w Context}}       &  & \multicolumn{3}{c|}{\textbf{w/o Context}}     & \multicolumn{3}{c|}{\textbf{w Context}}       \\ \cline{2-7} \cline{9-14} 
\multicolumn{1}{l|}{}                & \textbf{P}    & \textbf{R}    & \textbf{F1}   & \textbf{P}    & \textbf{R}    & \textbf{F1}   &  & \textbf{P}    & \textbf{R}    & \textbf{F1}   & \textbf{P}    & \textbf{R}    & \textbf{F1}   \\ \cline{1-7} \cline{9-14} 
\multicolumn{1}{|c|}{\textbf{T}}     & 68.8          & 68.8          & 68.8          & 68.4          & 68.4         & 68.4          &  & 74.3          & 74.2         & 74.2          & 72.8          & 72.8          & 72.8          \\
\multicolumn{1}{|c|}{\textbf{A}}     & 65.3          & 65.2          & 65.3          & 65.2          & 65.2          & 65.2          &  & 72.4          & 72.3          & 72.3          & 72.7          & 72.7          & 72.7          \\
\multicolumn{1}{|c|}{\textbf{V}}     & 68.7          & 68.4          & 68.4          & 68.0          & 68.0          & 68.0          &  & 70.5          & 70.5          & 70.5          & 70.5          & 70.5          & 70.5          \\
\multicolumn{1}{|c|}{\textbf{T+A}}   & \textbf{70.2} & \textbf{70.2} & \textbf{70.2} & \textbf{69.7} & \textbf{69.7} & \textbf{69.7} &  & 72.5          & 72.5          & 72.5          & 74.3         & 74.2          & 74.2          \\
\multicolumn{1}{|c|}{\textbf{A+V}}   & 71.1          & 71.1          & 71.1          & 71.3          & 71.2          & 71.2          &  & 71.8          & 71.6          & 71.5          & 71.0          & 70.9          & 70.9          \\
\multicolumn{1}{|c|}{\textbf{V+T}}   & 69.9          & 69.8          & 69.8          & 69.4          & 69.4          & 69.4          &  & \textbf{73.2} & \textbf{73.1} & \textbf{73.1} & 71.9          & 71.9          & 71.9          \\
\multicolumn{1}{|c|}{\textbf{T+A+V}} & 72.1          & 72.0          & 72.0          & 71.0          & 70.9          & 70.9          &  & 73.2          & 73.1          & 73.1          & \textbf{73.0} & \textbf{73.0} & \textbf{73.0} \\ \cline{1-7} \cline{9-14} 
\end{tabular}%
}
\caption{Sarcasm detection results for MUStARD, Weighted Average}
\label{tab:Sarcasm detection results for MUStARD}
\end{table*}


\subsection*{Additional Results}
\label{sec:additional_results}

As part of the bench-marking exercise, as mentioned in Section \ref{sec:experiments}, we perform Majority sampling (assigns the emotion class with majority examples as all samples), Random Sampling (predictions are sampled equally throughout the test set using this baseline). We also perform one-vs-rest experiments for each emotion which contain a sigmoid layer instead of a softmax layer as classification head. The results of these baselines are reported in the table \ref{tab:result-comparision}.
In order to study the advantage of our model, we presented a comparison of sarcasm detection results on original MUStARD dataset using our proposed model against the numbers reported by the authors of \citelanguageresource{mustard,ACL2020PB} in Table \ref{tab:sarcasm results comparison}. The detailed results of this experiment of sarcasm detection by using our proposed model on MUStARD across different modality, speaker, context combinations is given in Table \ref{tab:Sarcasm detection results for MUStARD}.

Table \ref{tab:Implicit-Emotion distribution in Sarcastic Utterances-ONEvsREST} and Table \ref{tab:explicit-Emotion distribution in Sarcastic Utterances-ONEvsREST} represent the detailed weighted average results of all modality combinations for ONEvsREST classification experiments in implicit and explicit emotions respectively.

Though the main focus of the paper is detection of emotion in sarcasm, in order to benchmark the entire data, we perform implicit and explicit emotion classification experiments in a multiclass setting. The results of these experiments across different modalities, context and speaker settings are presented in Table \ref{tab:implicit-emotion-in-mustard++} and Table \ref{tab:explicit-emotion-in-mustard++} for implicit and explicit emotions respectively.

\begin{table*}[ht!]
\centering
\resizebox{\textwidth}{!}{%
\begin{tabular}{c|ccc|ccc|l|ccc|ccc|}
\cline{2-14}
\multicolumn{1}{l|}{}                & \multicolumn{6}{c|}{\textbf{Speaker Independent}}                                             &  & \multicolumn{6}{c|}{\textbf{Speaker Dependent}}                                               \\ \cline{2-7} \cline{9-14} 
\multicolumn{1}{l|}{}                & \multicolumn{3}{c|}{\textbf{w/o Context}}     & \multicolumn{3}{c|}{\textbf{w Context}}       &  & \multicolumn{3}{c|}{\textbf{w/o Context}}     & \multicolumn{3}{c|}{\textbf{w Context}}       \\ \cline{2-7} \cline{9-14} 
\multicolumn{1}{l|}{}                & \textbf{P}    & \textbf{R}    & \textbf{F1}   & \textbf{P}    & \textbf{R}    & \textbf{F1}   &  & \textbf{P}    & \textbf{R}    & \textbf{F1}   & \textbf{P}    & \textbf{R}    & \textbf{F1}   \\ \cline{1-7} \cline{9-14} 
\multicolumn{1}{|c|}{\textbf{T}}    & 34.1 & 35.4 & 34.6 & 35.2 & 36.8 & 35.9 && 33.6 & 34.4 & 33.8 & \textbf{34.2} & \textbf{35.1} & \textbf{34.6}          \\
\multicolumn{1}{|c|}{\textbf{A}}     & 29.4          & 30.3          & 29.7          & 29.5          & 30.4          & 29.9          &  & 30.7          & 31.1          & 30.6          & 29.4          & 30.6          & 29.7          \\
\multicolumn{1}{|c|}{\textbf{V}}     & 31.7          & 34.6          & 32.9          & 32.7          & 31.3         & 31.8          &  & 32.2          & 33.6          & 32.5          & 31.1          & 32.8          & 31.8          \\
\multicolumn{1}{|c|}{\textbf{T+A}}   & \textbf{34.7} & \textbf{37.9} & \textbf{35.8} & 35.3 & 38.9 & 35.5 && 33.9 & 34.4 & 34.0 & 33.4 & 34.8 & 33.9          \\
\multicolumn{1}{|c|}{\textbf{A+V}}   & 29.8          & 31.6          & 30.5          & 29.3          & 33.4          & 30.6          &  & 31.8          & 33.4          & 32.5          & 30.2          & 32.6          & 30.6          \\
\multicolumn{1}{|c|}{\textbf{V+T}}  & 34.0 & 36.1 & 34.9 & \textbf{35.7} & \textbf{37.6} & \textbf{36.5} && \textbf{34.5} & \textbf{35.4} & \textbf{34.8} & 33.9 & 36.3 & 34.5          \\
\multicolumn{1}{|c|}{\textbf{T+A+V}} & 32.5 & 37.3 & 32.9 & 31.9 & 34.6 & 32.6 && 32.5 & 32.6 & 32.4 & 32.7 & 34.4 & 32.7
\\  \cline{1-7} \cline{9-14} 
 \hline
\end{tabular}%
}
\captionsetup{justification=centering}
\caption{Implicit Emotion Classification in Sarcastic Utterance (ONEvsREST) Weighted Average}
\label{tab:Implicit-Emotion distribution in Sarcastic Utterances-ONEvsREST}
\end{table*}

\begin{table*}[ht!]
    \centering
    \resizebox{\textwidth}{!}{%
        \begin{tabular}{c|ccc|ccc|l|ccc|ccc|}
            \cline{2-14}
            \multicolumn{1}{l|}{}                    & \multicolumn{6}{c|}{\textbf{Speaker Independent}} &                                          & \multicolumn{6}{c|}{\textbf{Speaker Dependent}}                                                                                                                                                                                                      \\ \cline{2-7} \cline{9-14}
            \multicolumn{1}{l|}{}                    & \multicolumn{3}{c|}{\textbf{w/o Context}}         & \multicolumn{3}{c|}{\textbf{w Context}}  &                                                 & \multicolumn{3}{c|}{\textbf{w/o Context}} & \multicolumn{3}{c|}{\textbf{w Context}}                                                                                                                                                                                                                                                                                                                 \\ \cline{2-7} \cline{9-14}
            \multicolumn{1}{l|}{}                    & \textbf{P}                                        & \textbf{R}                               & \textbf{F1}                                     & \textbf{P}                                & \textbf{R}                               & \textbf{F1}                              &  & \textbf{P}                               & \textbf{R}                               & \textbf{F1}                              & \textbf{P}                               & \textbf{R}                              & \textbf{F1}                              \\ \cline{1-7} \cline{9-14}
            \multicolumn{1}{|c|}{\textbf{T}}         & 39.8 & 42.8 & 41.0 & \textbf{ 40.2} & \textbf{43.6} & \textbf{41.5} && 37.6 & 41.6 & 38.5 & \textbf{39.4} & \textbf{41.6} & \textbf{40.3}                        \\
            \multicolumn{1}{|c|}{\textbf{A}}         & 28.3                                              & 31.8                                     & 29.3                                            & 27.7                                      & 30.8                                     & 28 8                                     &  & 29.8                                     & 33.4                                     & 31.3                                     & 30.5                                     & 35.9                                    & 31.6                                     \\
            \multicolumn{1}{|c|}{\textbf{V}}         & 26.6                                              & 30.6                                     & 28.2                                            & 28.3                                      & 30.1                                     & 29.0                                     &  & 29.3                                     & 33.6                                     & 30.9                                     & 29.0                                     & 33.4                                    & 30.6                                     \\
            \multicolumn{1}{|c|}{\textbf{T+A}}       & \textbf{41.1} & \textbf{43.9} & \textbf{42.2} & 39.7 & 41.9 & 40.5 && 38.1 & 40.6 & 39.1 & 37.5 & 39.4 & 38.3               \\
            \multicolumn{1}{|c|}{\textbf{A+V}}       & 29.6                                              & 31.9                                     & 30.4                                            & 27.7                                      & 30.0                                     & 28.6                                     &  & 30.1                                     & 34.3                                     & 31.2                                     & 29.8                                     & 31.9                                    & 30.7                                     \\
            \multicolumn{1}{|c|}{\textbf{V+T}}       & 41.2 & 43.6 & 42.1 & 39.9 & 41.9 & 40.7 && \textbf{38.8} & \textbf{40.4} &\textbf{ 39.5} & 36.6 & 38.4 & 37.4                             \\
            \multicolumn{1}{|c|}{\textbf{T+A+V}}     & 38.6 & 41.6 & 39.7 & 37.8 & 39.6 & 38.5 && 36.3 & 38.8 & 37.3 & 34.9 & 37.6 & 36.0              \\ \cline{1-7} \cline{9-14}
             \hline
        \end{tabular}%
    }
\captionsetup{justification=centering}
\caption{Explicit Emotion Classification in Sarcastic Utterance (ONEvsREST) Weighted Average}
\label{tab:explicit-Emotion distribution in Sarcastic Utterances-ONEvsREST}
\end{table*}

\begin{table*}[ht!]
    \centering
    \resizebox{\textwidth}{!}{%
        \begin{tabular}{c|ccc|ccc|l|ccc|ccc|}
            \cline{2-14}
            \multicolumn{1}{l|}{}                    & \multicolumn{6}{c|}{\textbf{Speaker Independent}} &                                          & \multicolumn{6}{c|}{\textbf{Speaker Dependent}}                                                                                                                                                                                                      \\ \cline{2-7} \cline{9-14}
            \multicolumn{1}{l|}{}                    & \multicolumn{3}{c|}{\textbf{w/o Context}}         & \multicolumn{3}{c|}{\textbf{w Context}}  &                                                 & \multicolumn{3}{c|}{\textbf{w/o Context}} & \multicolumn{3}{c|}{\textbf{w Context}}                                                                                                                                                                                                                                                                                                                 \\ \cline{2-7} \cline{9-14}
            \multicolumn{1}{l|}{}                    & \textbf{P}                                        & \textbf{R}                               & \textbf{F1}                                     & \textbf{P}                                & \textbf{R}                               & \textbf{F1}                              &  & \textbf{P}                               & \textbf{R}                               & \textbf{F1}                              & \textbf{P}                               & \textbf{R}                              & \textbf{F1}                              \\ \cline{1-7} \cline{9-14}
            \multicolumn{1}{|c|}{\textbf{T}}         & 22.6 & 23.6 & 23.0 & 23.3 & 24.3 & 23.6 && 25.7 & 25.9 & 25.1 & \textbf{25.4} & \textbf{26.3} & \textbf{25.3}                        \\
            \multicolumn{1}{|c|}{\textbf{A}}     &21.1 &22.3 &21.5 &20.6 &21.2 &20.7     & &20.7 &22.0 & 21.2 &21.1 &22.0 &21.5                                                             \\
            \multicolumn{1}{|c|}{\textbf{V}}         & 18.5                                              & 19.4                                     & 18.9                                            & 20.2                                      & 21.4                                     & 20.6                                     &  & 19.9                                     & 21.4                                     & 20.5                                     & 22.6                                     & 23.6                                    & 23                                     \\
            \multicolumn{1}{|c|}{\textbf{T+A}}       & \textbf{25.5} & \textbf{26.9} & \textbf{26.0} & 23.6 & 24.5 & 24.0  && \textbf{25.6} & \textbf{27.1} & \textbf{26.1} & 24.6 & 26.0 & 25.3               \\
            \multicolumn{1}{|c|}{\textbf{A+V}}       & 19.6                                              & 20.9                                     & 20.1                                            & 21.3                                      & 23.0                                     & 21.8                                     &  & 20.6                                     & 21.9                                     & 21.2                                     & 22.0                                    & 23.6                                    & 22.6                                     \\
            \multicolumn{1}{|c|}{\textbf{V+T}}       & 24.1 & 24.9 & 24.4 & 24.0 &25.3 & 24.6 && 23.0 & 24.4 &23.5 & 22.6 & 23.9 & 23.2                             \\
            \multicolumn{1}{|c|}{\textbf{T+A+V}}     & 23.4 & 24.7 & 23.8 & \textbf{25.7} & \textbf{26.9} & \textbf{26.1} && 25.6 & 26.9 & 25.9 & 24.3 & 25.5 & 24.7              \\ \cline{1-7} \cline{9-14}
             \hline
        \end{tabular}%
    }
\captionsetup{justification=centering}
\caption{Implicit Emotion Classification in MUStARD++ (Multiclass, Weighted Average)}
\label{tab:implicit-emotion-in-mustard++}
\end{table*}

\begin{table*}[ht!]
    \centering
    \resizebox{\textwidth}{!}{%
        \begin{tabular}{c|ccc|ccc|l|ccc|ccc|}
            \cline{2-14}
            \multicolumn{1}{l|}{}                    & \multicolumn{6}{c|}{\textbf{Speaker Independent}} &                                          & \multicolumn{6}{c|}{\textbf{Speaker Dependent}}                                                                                                                                                                                                      \\ \cline{2-7} \cline{9-14}
            \multicolumn{1}{l|}{}                    & \multicolumn{3}{c|}{\textbf{w/o Context}}         & \multicolumn{3}{c|}{\textbf{w Context}}  &                                                 & \multicolumn{3}{c|}{\textbf{w/o Context}} & \multicolumn{3}{c|}{\textbf{w Context}}                                                                                                                                                                                                                                                                                                                 \\ \cline{2-7} \cline{9-14}
            \multicolumn{1}{l|}{}                    & \textbf{P}                                        & \textbf{R}                               & \textbf{F1}                                     & \textbf{P}                                & \textbf{R}                               & \textbf{F1}                              &  & \textbf{P}                               & \textbf{R}                               & \textbf{F1}                              & \textbf{P}                               & \textbf{R}                              & \textbf{F1}                              \\ \cline{1-7} \cline{9-14}
            \multicolumn{1}{|c|}{\textbf{T}}         & \textbf{34.8} & \textbf{37.5} & \textbf{35.8} & 35.2& 39.5 & 36.5 && \textbf{35.5} & \textbf{36.7} & \textbf{35.7} & 35.2 & 37.3 & 35.8                        \\
            \multicolumn{1}{|c|}{\textbf{A}}         & 26.2                                              & 30.0                                    & 27.6                                            & 26.9                                      & 30.0                                     & 27.5                                     &  & 24.4                                     & 28.6                                     & 26.1                                     & 25.7                                     & 29.4                                    & 27.1                                     \\
            \multicolumn{1}{|c|}{\textbf{V}}         & 23.9                                              & 27.0                                     & 25.3                                            & 24.1                                      & 27.9                                     & 25.6                                     &  & 24.3                                     & 28.4                                     & 26.1                                     & 25.3                                     & 30.8                                    & 27.5                                     \\
            \multicolumn{1}{|c|}{\textbf{T+A}}       & 34.2 & 37.9 & 35.5 & \textbf{35.6} & \textbf{38.1} & \textbf{36.5} && 34.2 & 37.8 & 35.7 & 34.7 & 37.0 & 35.6               \\
            \multicolumn{1}{|c|}{\textbf{A+V}}       & 25.8                                             & 29.2                                    & 27.0                                            & 24.6                                      & 28.1                                     & 25.9                                     &  & 25.6                                     & 29.7                                     & 27.2                                     & 26.1                                     & 31.1                                    & 28.0                                     \\
            \multicolumn{1}{|c|}{\textbf{V+T}}       & 34.2 & 37.2 & 35.4 & 34.8 &37.9 & 36.1 && 33.3 & 36.8 &34.7 & \textbf{35.5} & \textbf{38.4} & \textbf{36.5}                             \\
            \multicolumn{1}{|c|}{\textbf{T+A+V}}     & 34.7 & 36.4 & 35.4 & 35.1 & 38.0 & 36.4 && 34.7 & 37.1 & 35.7 & 32.4 & 35.4 & 35.7              \\ \cline{1-7} \cline{9-14}
             \hline
        \end{tabular}%
    }
\captionsetup{justification=centering}
\caption{Explicit Emotion Classification in MUStARD++ (Multiclass, Weighted Average)}
\label{tab:explicit-emotion-in-mustard++}
\end{table*}

\end{document}